\documentclass[11pt, a4paper, logo, copyright]{googlecloud}

\pdftrailerid{redacted}

\makeatletter
\renewcommand\bibentry[1]{\nocitep{#1}{\frenchspacing\@nameuse{BR@r@#1\@extra@b@citeb}}}
\makeatother

\usepackage{kantlipsum, lipsum}
\usepackage{dsfont}
\usepackage[authoryear, sort&compress, round]{natbib}

\usepackage[utf8]{inputenc} %
\usepackage[T1]{fontenc}    %
\usepackage{hyperref}       %
\hypersetup{
  colorlinks,
  citecolor=blue,
  linkcolor=red,
  urlcolor=blue}
\usepackage{booktabs}       %
\usepackage{nicefrac}       %
\usepackage{microtype}      %
\usepackage{wrapfig} %
\usepackage{soul} %
\usepackage{tikz, adjustbox}
\usepackage{arydshln}
\usepackage[capitalize,noabbrev]{cleveref}
\usepackage{fancyvrb}

\usepackage{inconsolata}
\usepackage[skins,breakable,most]{tcolorbox}

\usepackage{caption}
\usepackage{subcaption}
\usepackage{enumitem}
\usepackage{flushend}
\usepackage{balance}
\usepackage{lineno}
\usepackage{amsmath,amssymb,amsfonts}
\usepackage{algorithm}
\usepackage{algpseudocode}
\usepackage{graphicx}
\usepackage{textcomp}
\usepackage{xcolor}
\usepackage{colortbl}
\usepackage{amsthm}
\usepackage{url}
\usepackage{float}
\usepackage{multirow}
\usepackage{multicol}
\usepackage{color}
\usepackage{bm}
\usepackage{bbm}

\usepackage{pifont}

\usepackage{array}
\newcolumntype{L}[1]{>{\raggedright\let\newline\\\arraybackslash\hspace{0pt}}m{#1}}
\newcolumntype{C}[1]{>{\centering\let\newline  \\\arraybackslash\hspace{0pt}}m{#1}}
\newcolumntype{R}[1]{>{\raggedleft\let\newline \\\arraybackslash\hspace{0pt}}m{#1}}

\usepackage[utf8]{inputenc} 
\usepackage[T1]{fontenc}    
\usepackage{amsfonts}       
\usepackage{comment} 
\usepackage{mdframed}
\usepackage{fontawesome}
\usepackage{csquotes}
\usepackage{makecell}
\definecolor{beigecolor}{RGB}{253, 244, 204} 
\definecolor{greencolor}{RGB}{228, 242, 217} 
\definecolor{bluecolor}{RGB}{66, 133, 244} 
\definecolor{orgcolor}{RGB}{255, 140, 15} 

\definecolor{redcolor}{RGB}{234, 67, 53} 

\definecolor{ggreen}{RGB}{52, 168, 83}

\definecolor{gyellow}{RGB}{251, 188, 5}

\lstdefinestyle{mystyle}{
    backgroundcolor=\color{backcolour},   
    commentstyle=\color{codegreen},
    keywordstyle=\color{magenta},
    numberstyle=\tiny\color{codegray},
    stringstyle=\color{codepurple},
    basicstyle=\ttfamily\scriptsize,
    breakatwhitespace=false,         
    breaklines=true,                 
    captionpos=b,                    
    keepspaces=true,                 
    numbers=left,                    
    numbersep=5pt,                  
    showspaces=false,                
    showstringspaces=false,
    showtabs=false,                  
    tabsize=2,
    frame=none,
    aboveskip=1pt,
    belowskip=1pt,
}
\lstdefinestyle{plainins}{
    backgroundcolor=\color{white},   
    commentstyle=\color{codegreen},
    keywordstyle=\color{magenta},
    numberstyle=\tiny\color{codegray},
    stringstyle=\color{codepurple},
    basicstyle=\ttfamily\scriptsize,
    breakatwhitespace=false,         
    breaklines=true,                 
    captionpos=b,                    
    keepspaces=true,                 
    numbers=none,                    
    numbersep=5pt,                  
    showspaces=false,                
    showstringspaces=false,
    showtabs=false,                  
    tabsize=2,
    aboveskip=0pt,
    belowskip=0pt,
    frame=single
}
\lstdefinestyle{plainexam}{
    backgroundcolor=\color[HTML]{FFFCF3},   
    commentstyle=\color{codegreen},
    keywordstyle=\color{magenta},
    numberstyle=\tiny\color{codegray},
    stringstyle=\color{codepurple},
    basicstyle=\ttfamily\scriptsize,
    breakatwhitespace=false,         
    breaklines=true,                 
    captionpos=b,                    
    keepspaces=true,                 
    numbers=none,                    
    numbersep=5pt,                  
    showspaces=false,                
    showstringspaces=false,
    showtabs=false,                  
    tabsize=2,
    aboveskip=0pt,
    belowskip=0pt
}

\tcbset{
  aibox/.style={
    width=\linewidth,
    top=5pt,
    bottom=1pt,
    colback=blue!6!white,
    colframe=black,
    colbacktitle=black,
    enhanced,
    center,
    fontupper=\scriptsize,
    attach boxed title to top left={yshift=-0.1in,xshift=0.15in},
    boxed title style={boxrule=0pt,colframe=white,},
  }
}
\newtcolorbox{AIbox}[2][]{aibox,title=#2,#1}
\definecolor{lightblue}{rgb}{0.22,0.45,0.70}

\title{Deep Researcher with Test-Time Diffusion}
\renewcommand{\today}{}

\author[1]{Rujun Han\textsuperscript{*}}
\author[1]{Yanfei Chen\textsuperscript{*}}
\author[2]{Zoey CuiZhu}
\author[1]{Lesly Miculicich}
\author[2]{Guan Sun}
\author[2]{Yuanjun Bi}
\author[2]{Weiming Wen}
\author[2]{Hui Wan}
\author[2]{Chunfeng Wen}
\author[2]{Solène Maître}
\author[1]{George Lee}
\author[2]{Vishy Tirumalashetty}
\author[2]{Emily Xue}
\author[2]{Zizhao Zhang}
\author[2]{Salem Haykal}
\author[1]{Burak Gokturk}
\author[1]{Tomas Pfister}
\author[1]{Chen-Yu Lee}

\affil[1]{Google Cloud AI Research}
\affil[2]{Google Cloud}

\begin{abstract}
Deep research agents, powered by Large Language Models (LLMs), are rapidly advancing; yet, their performance often plateaus when generating complex, long-form research reports using generic test-time scaling algorithms. Drawing inspiration from the iterative nature of human research, which involves cycles of searching, reasoning, and revision, we propose the Test-Time Diffusion Deep Researcher (TTD-DR). This novel framework conceptualizes research report generation as a diffusion process. TTD-DR initiates this process with a preliminary draft, an updatable skeleton that serves as an evolving foundation to guide the research direction. The draft is then iteratively refined through a "denoising" process, which is dynamically informed by a retrieval mechanism that incorporates external information at each step. The core process is further enhanced by a self-evolutionary algorithm applied to each component of the agentic workflow, ensuring the generation of high-quality context for the diffusion process. This draft-centric design makes the report writing process more timely and coherent while reducing information loss during the iterative search process. We demonstrate that our TTD-DR achieves state-of-the-art results on a wide array of benchmarks that require intensive search and multi-hop reasoning, significantly outperforming existing deep research agents.
\end{abstract}

\begin{document}
\maketitle
\section{Introduction}
Enabled by the recent advanced LLMs, building Deep Research (DR) agents has rapidly gained traction within both research and industry communities. These agents demonstrate remarkable capabilities, including the generation of novel ideas \citep{si2024canllm, hu2024nova}, effective information gathering through search tools \citep{Search-o1, jin2025search}, and the execution of analyses or experiments prior to drafting research reports or papers \citep{ai-scientist-v2, zheng-etal-2024-openresearcher}. 
\begin{wrapfigure}{r}{0.45\textwidth}
\vspace{-.5em}
\centering
\includegraphics[width=0.8\linewidth, trim=0cm 0cm 0cm 0cm]{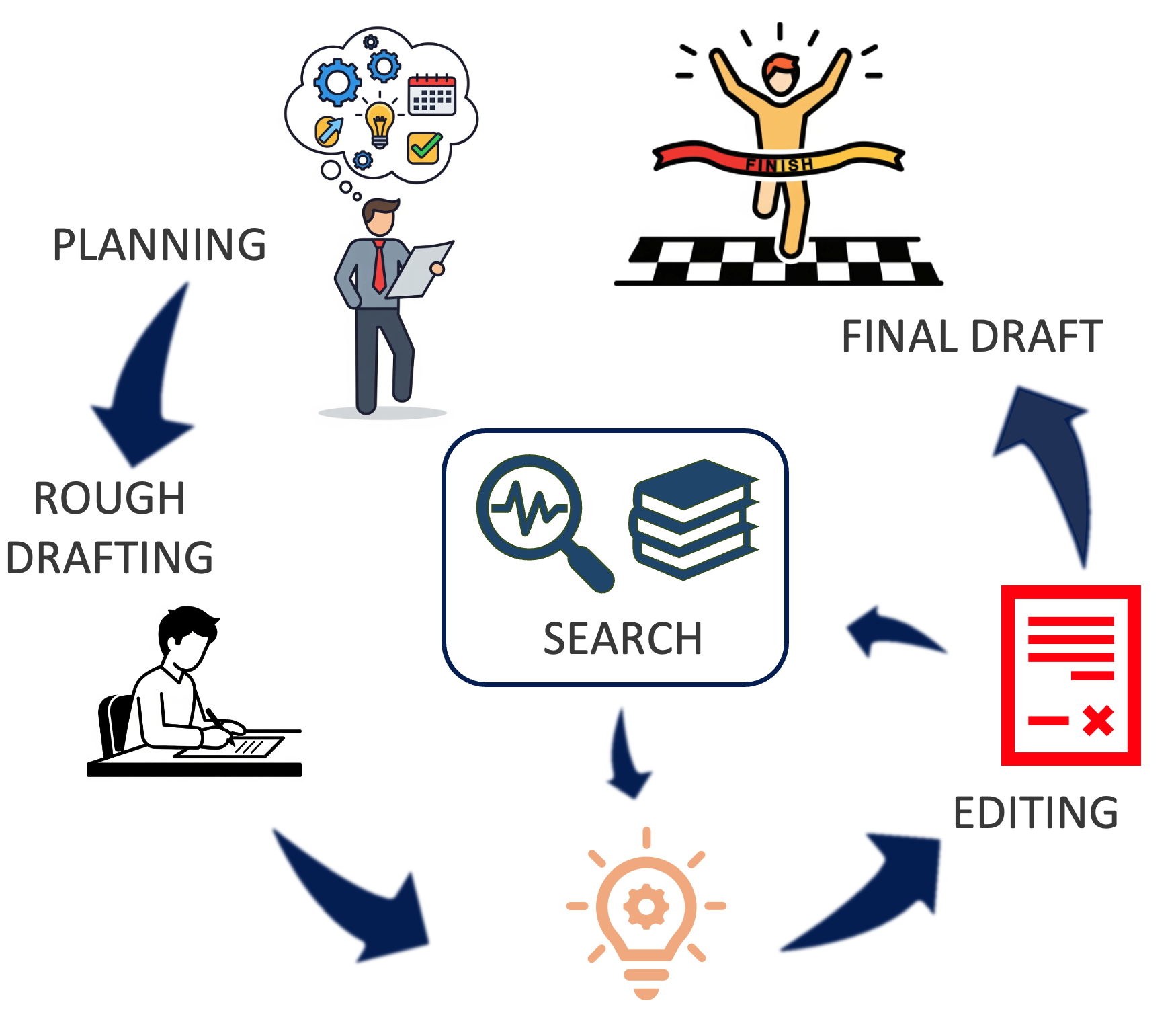}
\caption{Our method is inspired by the natural human writing process, which includes
planning, drafting, and multiple revisions to the draft.}
\label{fig:motivating}
\end{wrapfigure}
Existing DR agents primarily leverage test-time scaling approaches such as Chain-of-Thought (CoT) \citep{NEURIPS2022_9d560961}, best-of-n sampling\citep{ichihara2025evaluation}, Monte Carlo Tree Search \citep{MCTS}, debate mechanisms \citep{liang2023encouraging}, and self-refinement loops \citep{madaan2023selfrefine}. Despite the impressive progress, most popular public DR agents \citep{gpt-research, hf-deepresearch, open-deep-search} compile these test-time algorithms and various tools without a deliberate design driven by human cognitive behavior in writing, and commonly lack a principled draft, search, and feedback mechanism that empowers human researchers. This indicates a fundamental limitation in current DR agent work and highlights the need for a more cohesive, purpose-built framework for DR agents that imitates or surpasses human research capabilities.

Previous cognitive studies indicate that when human write about complex topics, they do not follow a linear progression, writing from the first word to the last. As Fig.~\ref{fig:motivating} \citep{writing-process} illustrates, people typically first establish a high-level \textit{plan}, then \textit{draft} the research report based on the plan, and subsequently engage in multiple \textit{revision} cycles \citep{writing-cognitive}. Crucially, during the revision phase, writers often seek out literature or search tools to gather supplementary information that refines and strengthens their arguments \citep{writing-information-seeking}.

\begin{figure}[t]
\centering
\includegraphics[scale=0.55, angle=-90, trim={7cm 0cm 7cm 0},clip]{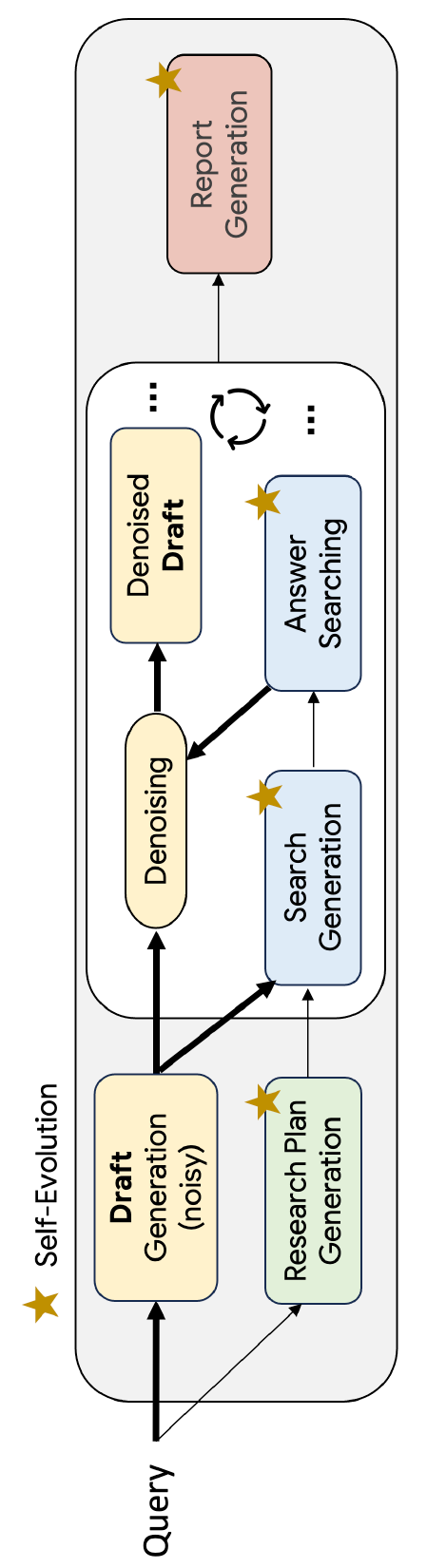}
    \vspace{-3mm}
	\caption{Illustration of our \textbf{Test-Time Diffusion Deep Researcher (TTD-DR)} framework, designed to mimic the iterative nature of human research through a \textit{draft}. A user query initiates both a preliminary draft and a research plan. This evolving draft, along with the research plan, dynamically informs the generation of search questions and subsequent information retrieval to be timely and coherent, while reducing information loss. The retrieved information is then leveraged to \textbf{denoise} and refine the initial draft in a continuous feedback loop. The entire workflow is further optimized by a self-evolutionary algorithm to enhance the quality of the research plan, generated questions, answers, and the final report, demonstrating the synergistic power of diffusion and self-evolution in achieving superior research outcomes.}
	\label{fig:intro}
\end{figure}

We observe a striking resemblance between this human writing pattern and the sampling process in a \textit{diffusion model} augmented by \textit{retrieval} \citep{ReDi}. In this analogy, a trained diffusion model initially generates a noisy draft, and the denoising module, aided by retrieval tools, revises this draft into higher-quality (or higher-resolution) outputs. Inspired by this diffusion sampling paradigm \citep{shen2025efficient, diffusionsurvey}, we propose \textbf{Test-Time Diffusion (TTD)} for deep research agents. Our framework meticulously models the entire research report generation as an iterative diffusion process, mirroring human cognitive patterns. As vanilla diffusion sampling can be ineffective for generating high quality outputs for complex research tasks, we specifically design our TTD Deep Researcher consisting of two mechanisms as illustrated by Fig.~\ref{fig:intro} and detailed below.

\textbf{(a) Denoising with Retrieval} \citep{ReDi}: An initial research report, drafted primarily from the LLM's internal knowledge, undergoes iterative refinement. The denoised draft, along with the research plan (Stage 1), guide the downstream research direction. Each denoising step is augmented by targeted retrieval of external information (Stage 2), significantly enhancing accuracy and comprehensiveness. \textbf{(b) Self-Evolution} \citep{lee-evolving, alpha-evolve}: Beyond the report-level diffusion through a draft, each individual component within the agentic workflow (e.g., plan, question, answer and report generation) undergoes its own optimization process. This encourages the exploration of diverse knowledge, mitigates the information loss for each unit agent throughout the long agentic trajectories, and thus provides more conducive context for report diffusion. The intricate interplay and synergistic combination of these two algorithms are crucial for achieving high quality research outcomes.

\begin{figure}[t]
	\begin{subfigure}[b]{0.43\linewidth}
	\captionsetup{justification=centering,margin=1cm}
		\centering
		\includegraphics[scale=0.25, angle=-90, trim={7cm 0.2cm 7cm 0},clip]{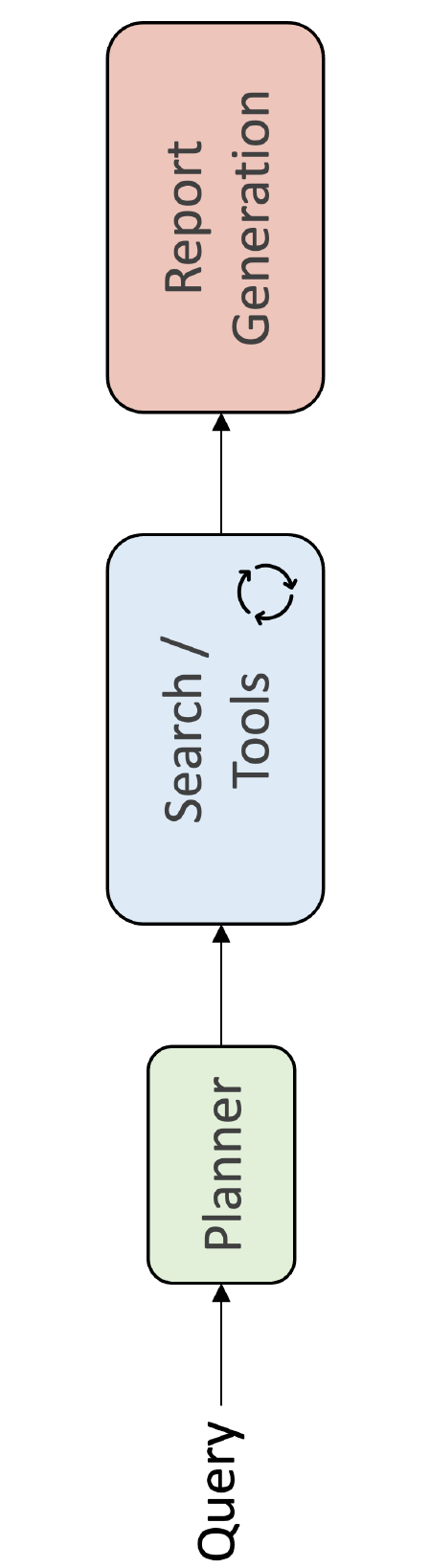}
		\caption{Huggingface Open DR}
		\label{fig:hf-dr}
	\end{subfigure}\qquad
	\begin{subfigure}[b]{0.47\linewidth}
	\captionsetup{justification=centering,margin=1cm}
		\centering
		\includegraphics[scale=0.28, angle=-90, trim={7cm 0.2cm 7cm 0},clip]{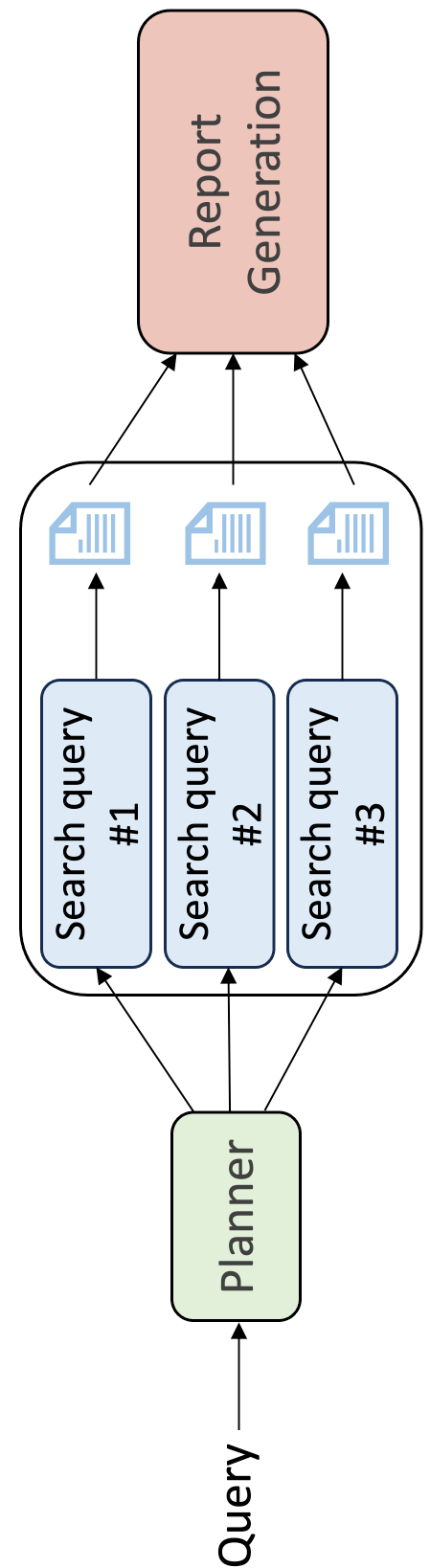}
		\caption{GPT Researcher}
		\label{fig:gpt-dr}
	\end{subfigure}
	\begin{subfigure}[b]{0.45\linewidth}
	\captionsetup{justification=centering,margin=1cm}
		\centering
		\vspace{3mm}
		\includegraphics[scale=0.28, angle=-90, trim={6cm 0cm 6cm 0},clip]{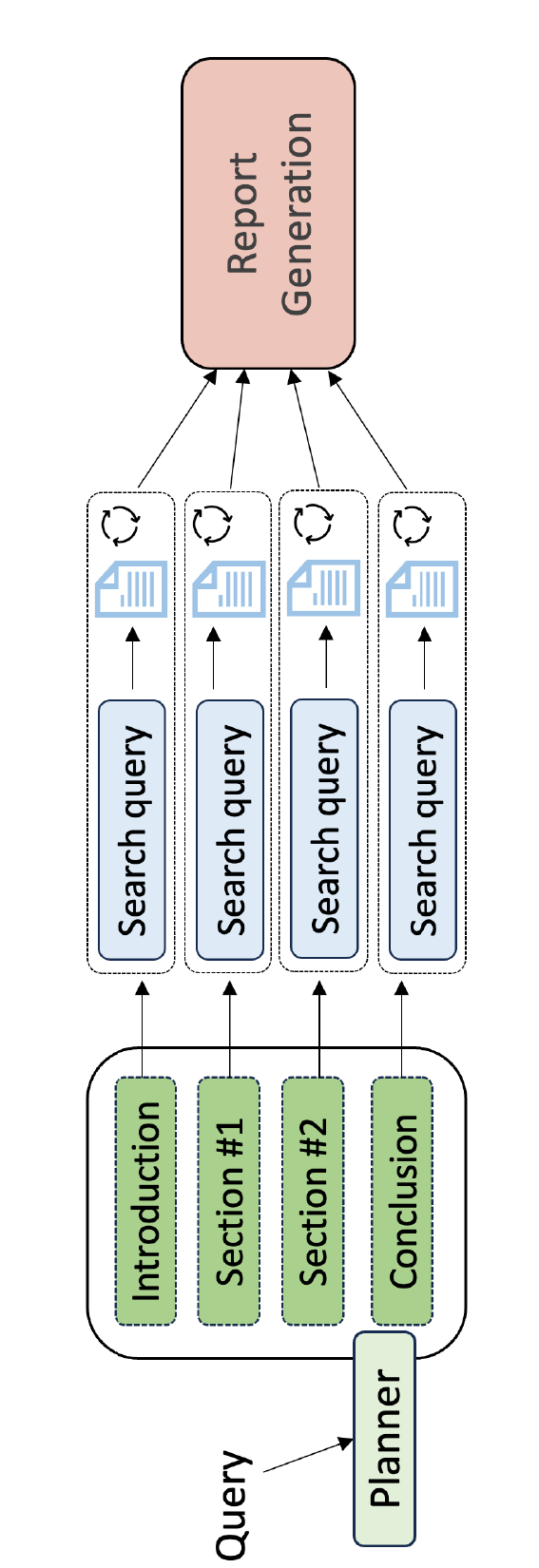}
		\vspace{-3mm}
		\caption{Open Deep Research}
		\label{fig:open-dr}
	\end{subfigure}\qquad
	\begin{subfigure}[b]{0.45\linewidth}
	\captionsetup{justification=centering,margin=1cm}
		\centering
		\vspace{3mm}
		\includegraphics[scale=0.3, angle=-90, trim={6cm 0cm 6cm 0},clip]{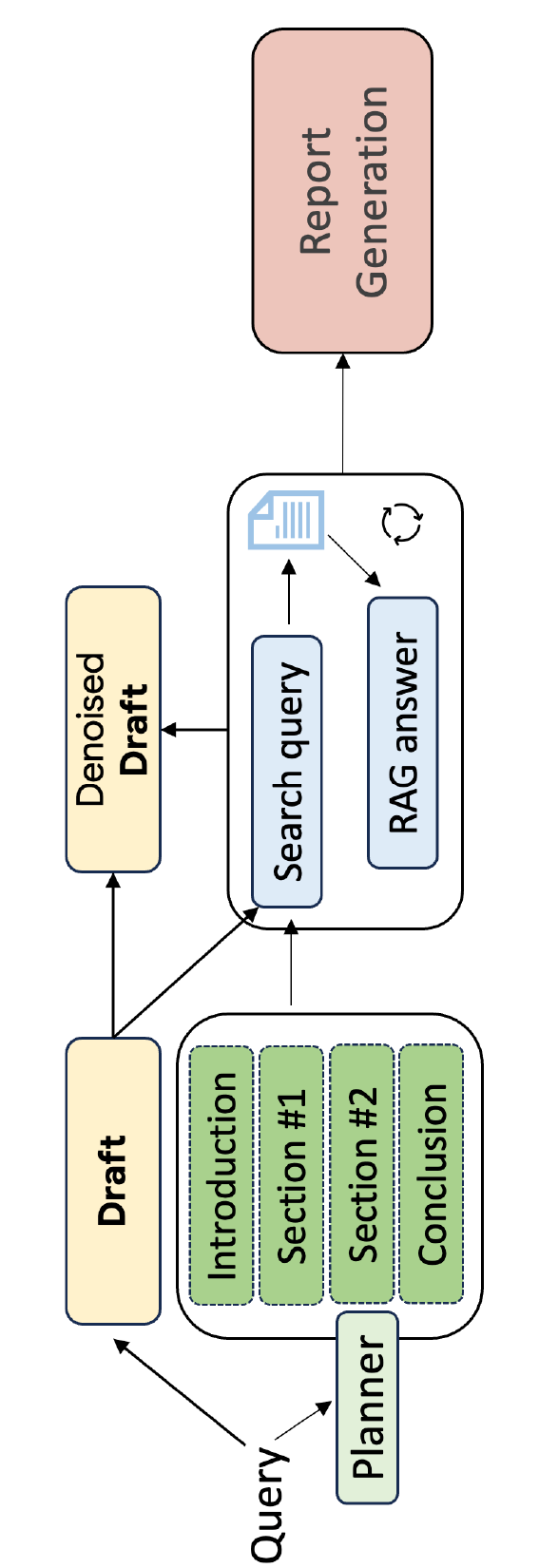}
		\vspace{-3mm}
		\caption{Test-Time Diffusion DR (ours)}
	\end{subfigure}
	\caption{A comparison of our method with other open-source deep researchers. (a) Huggingface Open DR \citep{hf-deepresearch} utilizes a lightweight planner to determine subsequent actions, such as calling search or browse tools, and repeats these actions until an answer is found. (b) GPT Researcher \citep{gpt-research} also employs a lightweight planner to generate and execute multiple search queries in parallel before a generator synthesizes the retrieved documents into a report. (c) Open Deep Research \citep{open-deep-research} uses a planner to outline the final report's structure and then conducts iterative research for each section individually before combining them. (d) Our \textbf{TTD-DR} introduces a draft denoising mechanism. Unlike Open Deep Research, TTD-DR avoids separated searches for each section to maintain global context and uses a RAG-based answer generator to process retrieved documents before saving them for the final report generation.}
	\label{comparision}
\end{figure}

Prior work primarily centers on scientific paper writing agents \citep{lu2024aiscientist, ai-scientist-v2, ai-coscientist, airesearcher, ai4research}, with a specific emphasis on generating academic publications. More recently, the scope has broadened to general research agents \citep{zhengdeep, webthinker} designed for broader information-seeking and reasoning use cases. In contrast to these existing efforts, our work introduce a deep research agent engineered for significantly broader applications. Specifically, we develop a research companion capable of generating helpful and comprehensive reports for complex research questions across diverse industry domains, including finance, biomedical, recreation, and technology \citep{han-etal-2024-rag}, similar to deep research products offered by \citet{OAIDeepResearch}, \citet{PerplexityDeepResearch} and \citet{grok}. Our framework targets search and reasoning-intensive user queries that current state-of-the-art LLMs cannot fully address using their internal knowledge or with conventional search tools. We summarize our key contributions below:
\begin{itemize}
    \vspace{-3mm}
    \item We propose a \textbf{Test-Time Diffusion Deep Researcher (TTD-DR)}, a novel test-time diffusion framework that enables the iterative drafting and revision of research reports, leading to more timely and coherent information integration while reducing information loss throughout the research process. 
    \item We stress test our \textbf{TTD-DR} using only search tools that are easily accessible to most agentic systems, eliminating the need to integrate additional proprietary tools (e.g., multimodal, web browsing).
    \item We establish a rigorous evaluation methodology for deep research agents, employing comprehensive metrics and expert evaluators. Our experiments demonstrate that \textbf{TTD-DR} substantially outperforms various leading research agents for tasks either require writing a long and comprehensive research report or need multi-hop search and reasoning to identify concise answers.
    \item We conduct a comprehensive ablation study and in-depth analysis to elucidate the individual contributions of \textbf{TTD-DR}'s components and demonstrate its effectiveness in surpassing leading DR agents.
\end{itemize}

\section{Test-Time Diffusion Deep Researcher (TTD-DR)}
\label{sec:methods}

Our approach, the Test-Time Diffusion Deep Researcher (TTD-DR), is inspired by the iterative nature of human research, which involves cycles of planning, drafting, searching for information, and revision. We conceptualize the generation of a complex research report as a \textit{diffusion} process where an initial, noisy draft is progressively refined into a high-quality final output. This is achieved through two core mechanisms operating in synergy: (1) Report-Level Refinement via \textbf{Denoising with Retrieval}, where the entire report draft evolves, and (2) Component-wise Optimization via \textbf{Self-Evolution}, which enhances the quality of each step in the research workflow.

The TTD-DR framework is designed to address the limitations of existing DR agents. As illustrated in Figure~\ref{comparision}, many public agents like Huggingface Open DR~\citep{hf-deepresearch}, GPT~\citep{gpt-research} Researcher , and Open Deep Research~\citep{open-deep-search} employ a linear or parallelized process of planning, searching, and generation. This can lead to a loss of global context and miss critical dependencies during the research process. Our draft-centric, iterative approach maintains coherence and provides a dynamic guide for the research direction, mitigating information loss. Proprietary DRs from \cite{OAIDeepResearch}, \cite{PerplexityDeepResearch} and \cite{grok} remain largely black box.

\subsection{Backbone Deep Research Agent}
\label{sec:dr-design}
\begin{figure}[h]
\centering
\includegraphics[scale=0.45, angle=-90, trim={8cm 0cm 8cm 0},clip]{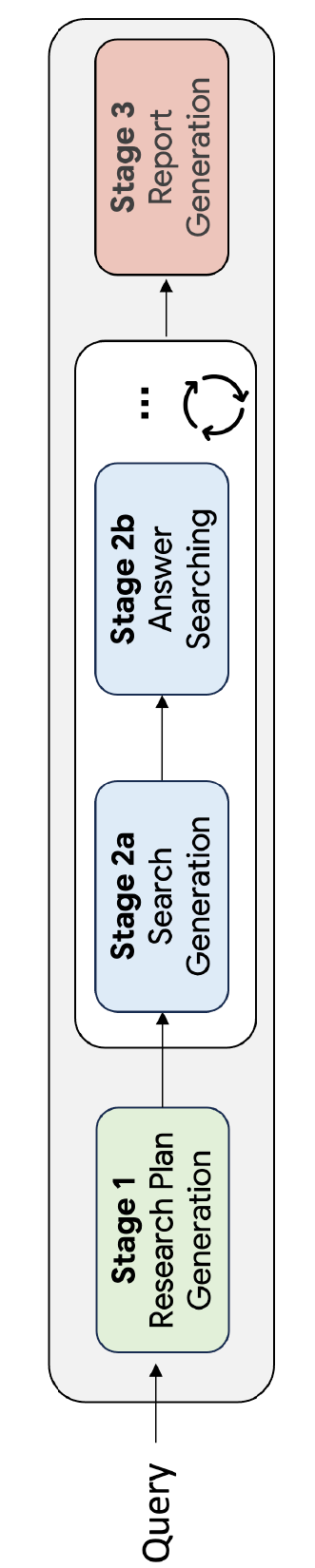}
\vspace{-4mm}
\caption{Our backbone DR agent operates in three stages, as illustrated above. \textbf{Stage 1} generates a detailed research plan that outlines the final report's structure and guides the information search. \textbf{Stage 2} iteratively generates search questions (2a) and then uses a RAG-like system to synthesize precise answers from retrieved documents (2b), rather than saving raw data. Finally, \textbf{Stage 3} synthesizes all gathered information to produce the final report. Each stage can be individually optimized using a self-evolutionary algorithm detailed in Sec.~\ref{self_evo}.}
\label{fig:dr-backbone}
\end{figure}

Fig.~\ref{fig:dr-backbone} illustrates our backbone deep research agent consisting of 3 major stages with several key components for an agentic framework: unit LLM agent, workflows and agent states. We explain them in details below.

\noindent
\textbf{Stage 1: Research Plan Generation} is a dedicated unit LLM agent which generates a structured research plan upon receiving a user query. This plan outlines a list of key areas needed for the final report, serving as an initial scaffold to guide the subsequent information-gathering process. Once a research plan is generated, it will be saved in agent stages and then transferred to its sub-agent.

\noindent
\textbf{Stage 2: Iterative Search and Synthesis} is a loop workflow nested in its parent sequential workflow. It contains of two sub-agents: \texttt{Search Question Generation (Stage 2a)} formulates a search query based on the research plan, the user query, and the context from previous search iterations (i.e., past questions and answers). \texttt{Answer Searching (Stage 2b)} searches the available sources (such as Google search) to find relevant documents and returns a summarized answer. This loop (\texttt{Stage 2a} $\rightarrow$ \texttt{Stage 2b}) continues until the research plan is adequately covered or a maximum number of iterations is reached.

\noindent
\textbf{Stage 3: Final Report Generation} is a unit LLM agent in its parent sequential workflow (\texttt{Stage 2} $\rightarrow$ \texttt{Stage 3}), which generates a comprehensive and coherent final report by synthesizing all the structured information gathered -- the plan from \texttt{Stage 1} and the series of question-answer pairs from \texttt{Stage 2}.

\subsection{Component-wise Self-Evolution}
\label{self_evo}
\begin{figure}[h]
\centering
\includegraphics[scale=0.55, angle=-90, trim={5.8cm 0cm 6cm 0},clip]{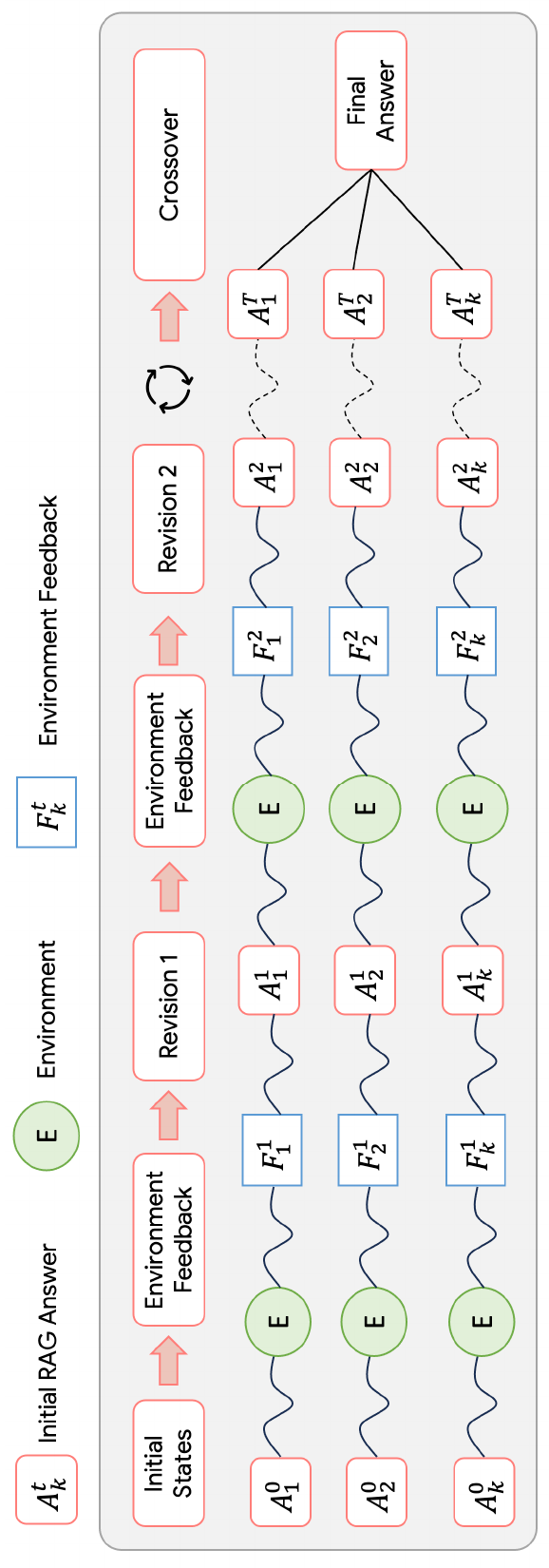}
\caption{Illustration of the component-wise \textbf{Self-Evolution} applied to Search Answer (Stage 2b in Figure~\ref{fig:dr-backbone}). The process starts with multiple variants of initial answers. Each variant then undergoes a self-evolving episode where it first interacts with the environment to obtain a fitness score and feedback. It is then revised based on the feedback. This process repeats until the maximum number of iterations is reached. Finally, multiple revised variants from all episodes are merged to produce the final answer.}
\label{fig:parallel-diffuse}
\end{figure}

The backbone DR agent introduced above determines the overall research directions (\texttt{Stage 1}), and supplies the context and information (\texttt{Stage 2}) for the final report writing (\texttt{Stage 3}). We enhance the performance of each stage's agents in order to \textit{find} and \textit{preserve} the high quality context. To accomplish this goal, we leverage a self-evolutionary algorithm to improve each stage's agents. Figure~\ref{fig:parallel-diffuse} illustrates our proposed algorithm inspired by recent self-evolution work \citep{lee-evolving, alpha-evolve}. Here we use the search answer generation as an example, but this algorithm can be applied to all stage agents such as plan generation, search question and even the final report generation to improve their output quality. This algorithm is implemented in a parallel workflow with the following sequential and loop workflows.

\begin{enumerate}
    \item \textbf{Initial States}. The leftmost blocks produce multiple diverse variants of an output (e.g., several possible answers to a search query) conditioned on the output of previous stages. Each block is implemented with a unit LLM Agent, allowing for the sampling of multiple answers using varied parameters (e.g., temperature, top\_k) to explore a larger search space. This ideally leads to discovery of potentially more valuable information.
    \item \textbf{Environmental Feedback}. Each answer variant is assessed by an LLM-as-a-judge, utilizing auto-raters for metrics such as Helpfulness and Comprehensiveness. These raters not only provide fitness scores but also generate textual critiques that help improve the answer. 
    \item \textbf{Revision Step}. With the scores and feedback from the previous step, each variant undergoes a revision step to adapt toward better fitness scores. The ``Environmental Feedback'' and ``Revision'' steps repeat until a stopping criterion is met, forming a loop workflow.
    \item \textbf{Cross-over}. Finally, multiple revised variants are merged into a single, high-quality output. This merging process consolidates the best information from all evolutionary paths, producing superior context for the main report generation process. The merging prompt can be found in Appendix~\ref{sec:report-merge}.
\end{enumerate}

While self-evolution improves the quality of each component's output, this information is not incorporated into the final report until the search process is complete. This delay motivates our second mechanism, Denoising with Retrieval, which integrates the agent's findings in a more timely and coherent manner to guide the research direction effectively.

\subsection{Report-level Denoising with Retrieval}

Inspired by the sampling process in diffusion models, where a noisy image is iteratively refined, we prompt an LLM to generate an initial draft report based on the user's query. This draft serves as a ``noisy'' starting point, as illustrated in Figure~\ref{fig:intro}. However, as noted in prior work, having a model denoise its own output without external context can lead to slow convergence and sub-optimal results \citep{ReDi, shen2025efficient, fast-monte-carlo}. This is particularly true for complex research queries where external information from search tools is essential for improving the draft. This observation motivates us to design a retrieval-augmented denoising process connected directly to our backbone DR workflow introduced in Sec.~\ref{sec:dr-design}.

Specifically, as shown in Algorithm~\ref{alg:denoising}, we feed the current draft report into \texttt{Stage 2a} of the backbone DR workflow to inform the generation of the next search query (Line 2). After obtaining a synthesized answer in \texttt{Stage 2b} (Line 4), the new information is used to revise the report draft, either by adding new details or by verifying existing information (Line 6). This process—feeding the denoised report back to generate the next search query—is repeated in a continuous loop. The draft is progressively "denoised" until the search process concludes, at which point a final agent writes the final report based on all historical search answers and revisions (\texttt{Stage 3}).

\begin{algorithm}[h]
\scriptsize
\caption{Denoising with Retrieval}\label{alg:denoising}
\begin{algorithmic}[1]
\Require $q$, $\mathcal{M}$, $\mathcal{P}$, $\mathcal{R}_0$, $\mathcal{Q}$, $\mathcal{A}$ \Comment{query, all agents, plan, initial noisy draft, history of search questions and answers}
\For{$t \in \{1,\dots,N\}$} \Comment{$N$: max number of revision steps}
\State $Q_t = \mathcal{M}_{\mathcal{Q}}(q, \mathcal{P}, \mathcal{R}_{t-1},  \mathcal{Q}, \mathcal{A}$) \Comment{generate the next question to address gaps in $\mathcal{R}_t$}
\State $Q_t \rightarrow \mathcal{Q}$ 
\State $A_t = \mathcal{M}_{\mathcal{A}}(Q_t)$ \Comment{retrieve external information to provide concrete delta for denoising}
\State $A_t \rightarrow \mathcal{A}$ 
\State $\mathcal{R}_t = \mathcal{M}_{\mathcal{R}}(q, \mathcal{R}_{t-1}, \mathcal{Q}, \mathcal{A}$) \Comment{remove ``noise'' (imprecision, incompleteness) from the previous draft}
\If{exit\_loop} 
\State break \Comment{if exit\_loop is called, stop revision}
\EndIf
\EndFor
\end{algorithmic}
\end{algorithm}

In summary, this continuous feedback loop, where the evolving draft guides the search and the search refines the draft, ensures the report remains coherent and the research stays on track. The final, "denoised" report is generated after the search process concludes, based on the full history of revisions and retrieved answers. The synergy between the component-wise self-evolution and the report-level diffusion process is critical, allowing TTD-DR to achieve state-of-the-art results.
\section{Experimental Setup}
\label{sec:eval}
To rigorously evaluate our Test-Time Diffusion Deep Researcher (TTD-DR), we established a comprehensive experimental framework. This section details the evaluation metrics, the datasets used for benchmarking, and the specifics of our implementation.

\subsection{Evaluation Metrics}

Our DR agent is inherently a complicated multi-agent system. Each stage of this system generates long responses that the final agent combine coherently to produce a comprehensive report for users. Evaluating long-form LLM responses and complex agentic trajectories presents significant challenges due to the vast number of facts to verify, intricate long-term logical dependencies, and the inherent subjectivity of both LLM and human judges \citep{han-etal-2024-rag, si2024canllm, li2024llmasajudge}. To ensure quality and efficiency of our evaluators, we collect high-quality human judgment annotations, calibrate LLM-as-a-judge calibrated with human preferences, and use the calibrated LLM-as-a-judge as the final evaluator. We provide more details of evaluation metrics below.

\begin{itemize}
    \vspace{-3mm}
    \item \textbf{Helpfulness} and \textbf{Comprehensiveness} are the two most commonly used metrics for evaluating long-form LLM responses, particularly for research outputs \citep{deepconsult, deepresearchgym, schmidgall2025agentlaboratoryusingllm}. We therefore adopt these two metrics and construct a new side-by-side quality comparison framework based on them. \textbf{Helpfulness} is defined by four criteria: 1) satisfying user intent, 2) ease of understanding (fluency and coherence), 3) accuracy, and 4) appropriate language. \textbf{Comprehensiveness} is defined as the absence of missing key information. Web search is permitted to better understand the query if needed. Guidelines for determining the Helpfulness and Comprehensiveness levels of a report can be found in Appendix~\ref{sec:eval-guideline}.
    \vspace{1mm}
    \item \textbf{Side-by-side quality comparison} (also known as pairwise evaluation), a widely adopted method for assessing long-form LLM responses \citep{han-etal-2024-rag, li2024llmasajudge, si2024canllm, liu2024aligning}. Raters were asked to express their preference between two reports (A and B) considering both Helpfulness and Comprehensiveness, using the following scale: 1) \textbf{Much Better} If A is both more helpful and more comprehensive than B; 2) \textbf{Better} If A is more helpful than B and equally comprehensive as B, or if A is more comprehensive than B and equally helpful as B; 3) \textbf{Slightly Better} If A is more helpful but less comprehensive than B; Otherwise, select 4) \textbf{About The Same} If none of the above conditions are met. The same logic applies when B is better than A. Our custom-built human annotation interface can be found in Appendix~\ref{sec:eval-interface}. Each pair is scored twice to compute agreement among human raters. We then deploy an LLM-as-a-judge with the same human instructions to align with human ratings. We discuss more calibration details in the next subsection.
    \vspace{1mm}
    \item \textbf{Correctness} is used for our multi-hop short-form QA tasks \citep{phan2025humanitysexam}. For such tasks, we can simply prompt LLMs to compare the long-form answers produced by our agents with the given ground-truths. We follow the standard evaluation prompt\footnote{https://scale.com/leaderboard/humanitys\_last\_exam\_text\_only} to first extract a single answer from LLMs' responses and then compare the extracted answers with ground-truths.
\end{itemize}

\begin{figure}[t]
\centering
	\begin{subfigure}[b]{0.42\linewidth}
	\captionsetup{justification=centering,margin=1cm}
		\centering
		\includegraphics[trim={0 0cm 0 0},clip, width=\textwidth]{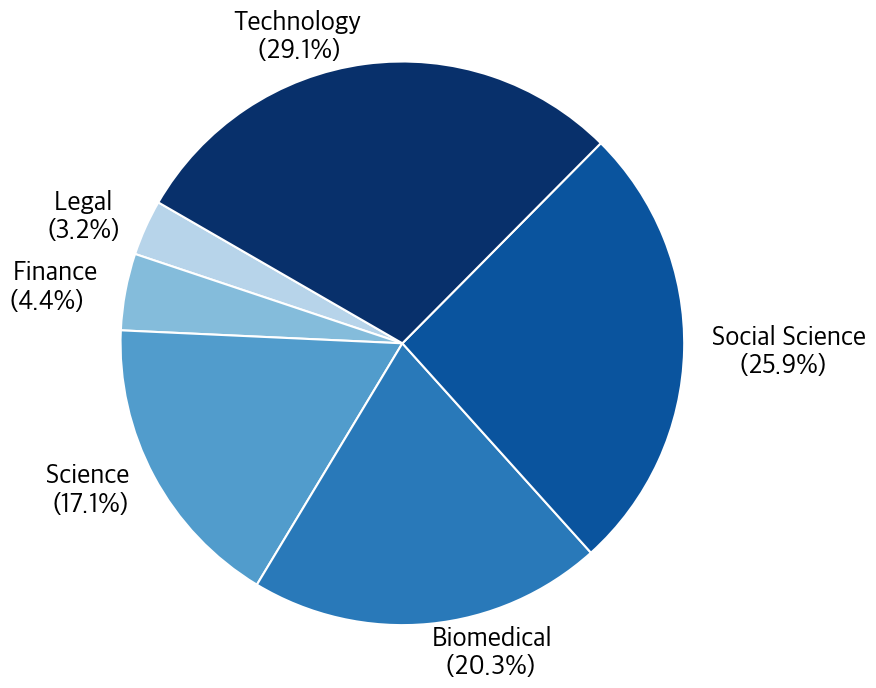}
		\caption{LongForm Research}
	\end{subfigure}\qquad
	\begin{subfigure}[b]{0.42\linewidth}
	\captionsetup{justification=centering,margin=1cm}
		\centering
		\includegraphics[trim={0 0cm 0 0},clip, width=\textwidth]{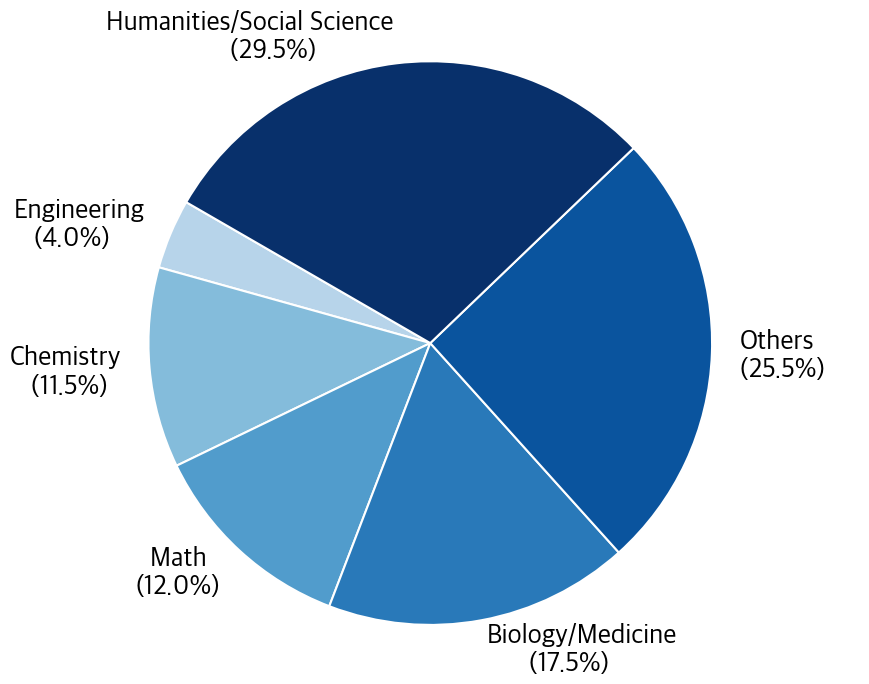}
		\caption{HLE-search}
	\end{subfigure}
	\caption{Query domain distribution of the evaluation sets: \textsc{LongForm Research} (left) and \textsc{HLE-search} (right), both demonstrating diverse domain coverage.}
	\label{fig:data-dist}
\end{figure}

\subsection{LLM-as-a-judge Calibration}

Given the absence of ground truth for long-form responses in the \textsc{LongForm Research} and \textsc{DeepConsult} benchmarks, a common practice for scalable evaluation is to leverage LLM-as-a-judge \citep{han-etal-2024-rag, si2024canllm, deepconsult, deepresearchgym, schmidgall2025agentlaboratoryusingllm}. However, most prior work in DR agents has not specifically calibrated LLM-as-a-judge's quality with human raters, raising questions regarding the reliability of auto-evaluators.

In contrast, we align our LLM-as-a-judge with human ratings by comparing 200 reports from our DR agents with those from OpenAI Deep Research. We then utilize an evaluator prompt similar to the one used in our human evaluation for side-by-side comparisons and then calculate the alignment scores between the auto-raters and human raters.  Table~\ref{tab:results-align} in Appendix~\ref{sec:human-llm-align} provides details and results regarding our selection of Gemini-1.5-pro as our LLM-as-a-judge.

For the Correctness auto-rater used to assess the \textsc{HLE} and \textsc{GAIA} dataset, we do not calibrate it with human ratings. This is because an official evaluation prompt exists for these tasks, and we maintain consistency with the research community by adhering to the original prompt. Furthermore, all answers in these two benchmarks have a straightforward ground-truth answer, simplifying the judgment of LLM response correctness. Therefore, we use Gemini-1.5-pro as the evaluator model without further human calibration for these specific tasks.

\subsection{Data}
Our chosen benchmarks focus on two broad tasks. 1) Complex queries that require research agents to produce a long-form comprehensive report (LongForm Research and DeepConsult) 2) multi-hop queries that require extensive search and reasoning to answer (HLE and GAIA). Both categories fit into our objective of building a general-purpose, real-world research companion, similar to \textsc{OpenAI Deep Research} \citep{OAIDeepResearch} and \textsc{Perplexity Deep Research} \citep{PerplexityDeepResearch}. Notably, both tasks may require up to 20 search steps (hops) to fully address user queries, as show in Figure~\ref{fig:pareto-frontier-revision} and \ref{fig:pareto-frontier-revision-hle} in the appendix. Other datasets are outside the scope of this work if they do not require extensive search (e.g., only need a few search steps), such as long-form RAG-QA \citep{han-etal-2024-rag, stelmakh-etal-2022-asqa} and short-form multi-hop QA \citep{yang-etal-2018-hotpotqa, trivedi-etal-2022-musique}. This also applies to datasets not targeting general-purpose research report generation, such as AI-Researcher \citep{airesearcher}. Additionally, we focus on search tool usage, deferring the incorporating of other tools such as browsing and coding to future work.

\noindent
\textbf{LongForm Research.} To benchmark our DR agent system against other baselines, we first curate a set of licensed real-world queries that demand search and complex reasoning. This dataset best represents our target use cases where users require deep research to create helpful and comprehensive reports. This evaluation set comprises 205 queries covering multiple industry domains, as demonstrated in Figure.~\ref{fig:data-dist}.

\noindent
\textbf{DeepConsult} \citep{deepconsult} is a collection of business and consulting-related prompts designed for deep research. The query set spans a wide range of topics, including marketing, finance, technology trend and business planning.

\noindent
\textbf{Humanity's Last Exam (HLE)} \citep{phan2025humanitysexam} is a benchmark of 2,500 extremely challenging questions across dozens of subject areas, intended as the final closed-ended benchmark for broad academic capabilities. We focus on the text-only subset, reserving the multi-modality for future research. We name this dataset \textsc{HLE-full}.

\noindent
\textbf{HLE-search}. A significant number of queries in the HLE dataset do not require extensive searching to resolve. To better benchmark our target use cases of search with reasoning, we identify queries from HLE that demand the most search capabilities. Specifically, we prompt the Gemini-1.5-pro model to categorize all queries into either [a] pure reasoning and [b] requiring search. The prompt used can be found in the Appendix~\ref{sec:query-cat}. Finally, we randomly sample 200 queries from categories [b]. As shown in  Table~\ref{tab:results-ablation}, the LLM's own performances on this curated subset is considerably lower compared with the full set. Its question domain distribution can also be found in Figure~\ref{fig:data-dist}. Therefore, we believe \textsc{HLE-search} serves as a more suitable benchmark for our research focus.

\noindent
\textbf{GAIA} \citep{gaia} is another public benchmark that evaluates AI on real-world questions, encompassing questions across three levels of difficulty. Successful completion of these tasks requires abilities such as reasoning, multi-modal fluency, web browsing, and tool-use proficiency. We use the evaluation set to compare with other baselines.

\subsection{Implementation Details}
\paragraph{Agentic Framework.} To implement our \textbf{TTD-DR}, we require a modular and easily extensible agent system capable of leveraging leading LLMs, such as Gemini-2.5-pro, to seamlessly orchestrate workflows, call tools, and execute tasks. Google Agent Development Kit (ADK)\footnote{https://google.github.io/adk-docs/} is a recently released agent development platform that satisfies all these requirements. All components described in Sec.~\ref{sec:methods} can be easily implemented with ADK. We thus chose to build our deep researcher based on ADK.

We fix maximum denoising with retrieval steps to 20. Other hyper-parameters for \textsc{Self-Evolution} algorithm can be found in Appendix~\ref{sec:hyper}. We use grounding with Google search\footnote{https://cloud.google.com/vertex-ai/generative-ai/docs/grounding/overview} to implement the RAG system in Stage 2b.

\begin{table*}[t]
\centering
\caption{In this table, we show our \textbf{TTD-DR}'s performances against different baseline systems for \textsc{LongForm Research}, \textsc{DeepConsult}, \textsc{HLE} and \textsc{GAIA} datasets. Win rate (\%) are computed based on OpenAI Deep Research. Correctness is computed as matching between system predicted and reference answers. For Grok DeeperSearch on \textsc{HLE-full}, there is no public number provided, and we are not able to scrape the full 2K queries due to research budget and Grok DeeperSearch's daily scrape limits.}
\resizebox{\textwidth}{!}{
\begin{tabular}{lccccc}\toprule[1.5pt]
& \textsc{LongForm Research} & \textsc{DeepConsult} & \textsc{HLE-Search} & \textsc{HLE-Full} & \textsc{GAIA} \\
\cmidrule{2-6}
& \textbf{Win Rate} & \textbf{Win Rate} & \textbf{Correctness} & \textbf{Correctness} & \textbf{Correctness} \\
\midrule
\textsc{OpenAI Deep Research} & - & - & 29.1 & 26.6 & 67.4 \\
\textsc{Perplexity Deep Research} & 21.8 & 32.0  & 14.5 & 21.1 & 54.5 \\
\textsc{Grok DeeperSearch} & 16.1 &  16.0 & 19.3 & - & 47.9 \\
\textsc{GPT-Researcher} & 18.3 & \phantom{0}9.4 & \phantom{0}2.0 & \phantom{0}4.1 & 37.7 \\
\textsc{Open Deep Search} & \phantom{0}2.6 & \phantom{0}2.2 & \phantom{0}3.0 & \phantom{0}0.4  & 20.9  \\
\midrule
\textbf{TTD-DR} (ours) & \textbf{69.1} & \textbf{74.5} & \textbf{33.9} & \textbf{34.3} & \textbf{69.1} \\
\bottomrule
\end{tabular}
}
\label{tab:results-main}
\end{table*}

\subsection{Compared Systems}
We compare our RA systems with the leading RA agents in the market: \textsc{OpenAI Deep Research} \citep{OAIDeepResearch}, \textsc{Perplexity Deep Research} \citep{PerplexityDeepResearch}, \textsc{Grok DeepSearch} \citep{grok}, \textsc{Open Deep Search} \citep{open-deep-search} and \textsc{GPT-Researcher} \citep{gpt-research}. For DR agents not supported by an API, we manually scraped and saved their raw outputs.

For ablation study, we compare with baseline LLMs Gemini-2.5-pro and Gemini-2.5-flash, along with their variants that include a simple search tool (simple RAG). For our DR Agent, we compare the following. 1) \textsc{Backbone DR Agent} is our backbone DR Agent without any test-time scaling algorithms. 2) \textsc{+Self-evolution} and 3) \textsc{+Denoising with retrieval} are two DR agent variants enhanced by our proposed test-time scaling algorithms. Our DR agents use Gemini-2.5-pro as the base model. All other baselines agents use their default LLMs (e.g. o3 for OpenAI DR).

\section{Results and Analysis}

\begin{figure}[t]
\centering
	\begin{subfigure}[b]{0.47\linewidth}
	    \captionsetup{justification=centering,margin=0cm}
		\centering
		\includegraphics[width=0.97\textwidth]{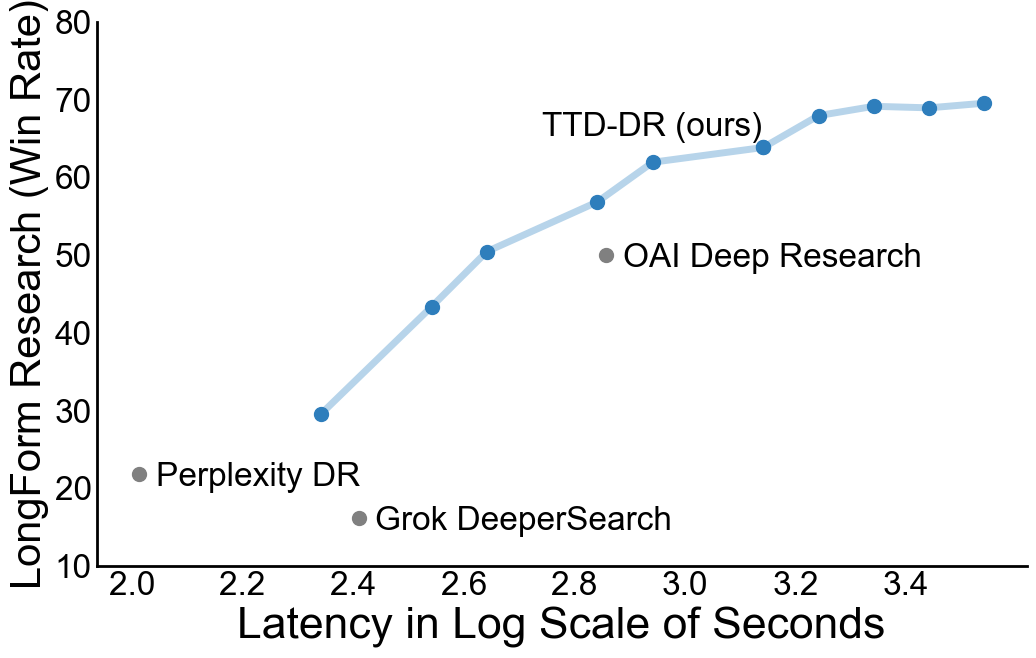}
	    \caption{Pareto frontier for revision steps.}
	    \label{fig:pareto-frontier-revision}
	\end{subfigure}\qquad
	\begin{subfigure}[b]{0.47\linewidth}
	    \captionsetup{justification=centering,margin=0cm}
		\centering
		\includegraphics[width=0.97\textwidth]{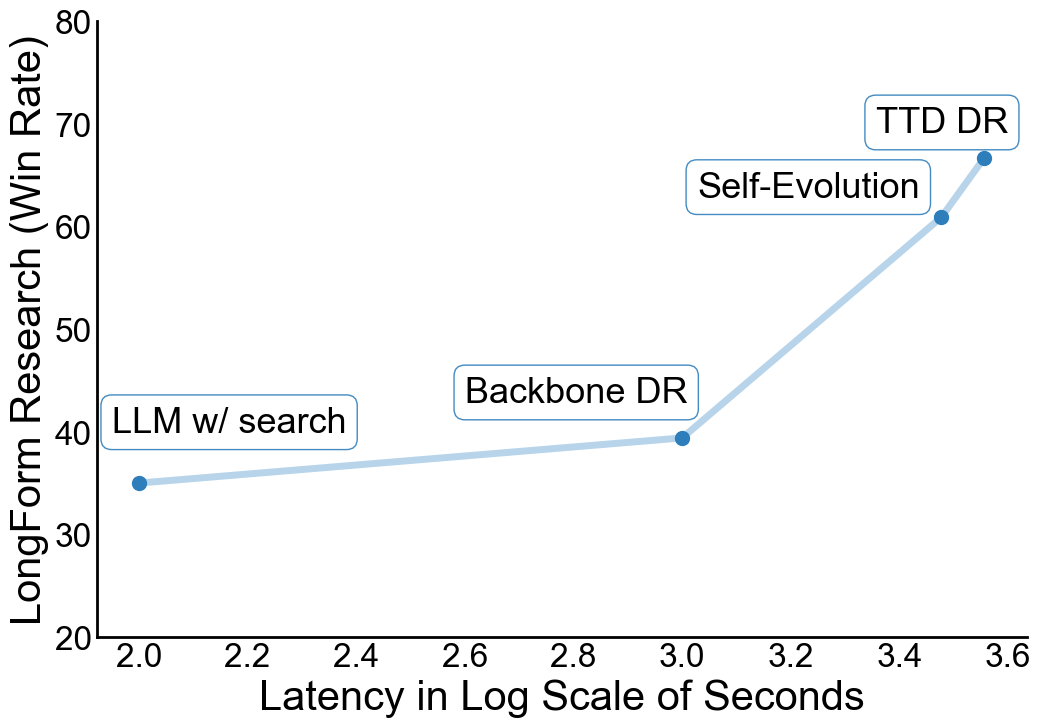}
	    \caption{Pareto frontier for different DR designs.}
	    \label{fig:pareto-frontier}
	\end{subfigure}
	\caption{Pareto frontier between DR agent performances and latency for \textsc{LongForm Research}. \textbf{Left}: the dots from left to right represent adding more search/revision steps up to 20, which shows with similar latency, we achieve better or on-par compared with other DR agents. \textbf{Right}: the dots from left to right represent 1) \textsc{Gemini-2.5-pro w/ search tool}, 2) \textsc{Backbone DR Agent}, 3) \textsc{+Self-evolution} and 4) \textsc{+Diffusion with Retrieval}, which shows our final algorithm is most efficient in terms of test-time scaling (steepest slope).}
\end{figure}

\subsection{Main Results}
\begin{figure}[t]
\centering
	\begin{subfigure}[b]{0.42\linewidth}
	\captionsetup{justification=centering,margin=1cm}
		\centering
		\includegraphics[width=\textwidth]{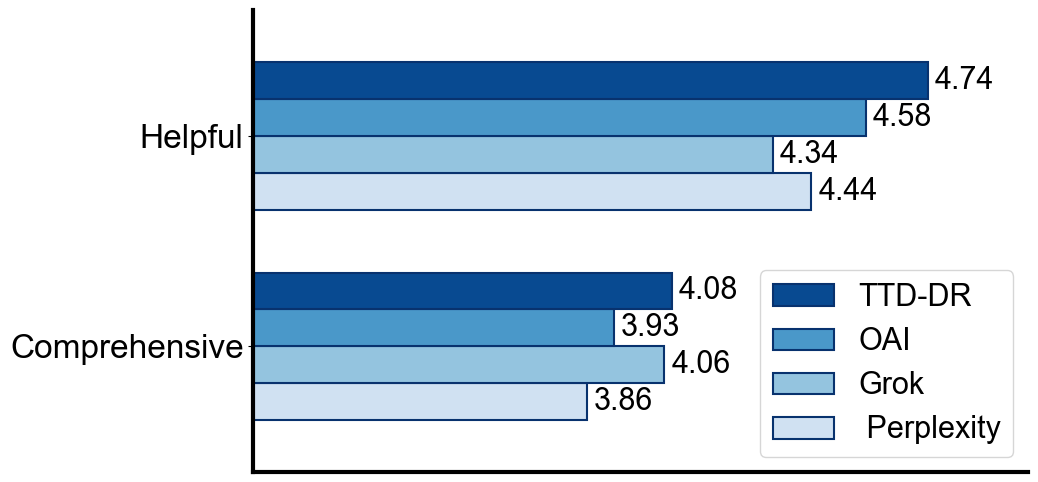}
		\caption{LongForm Research}
	\end{subfigure}\qquad
	\begin{subfigure}[b]{0.42\linewidth}
	\captionsetup{justification=centering,margin=1cm}
		\centering
		\includegraphics[width=\textwidth]{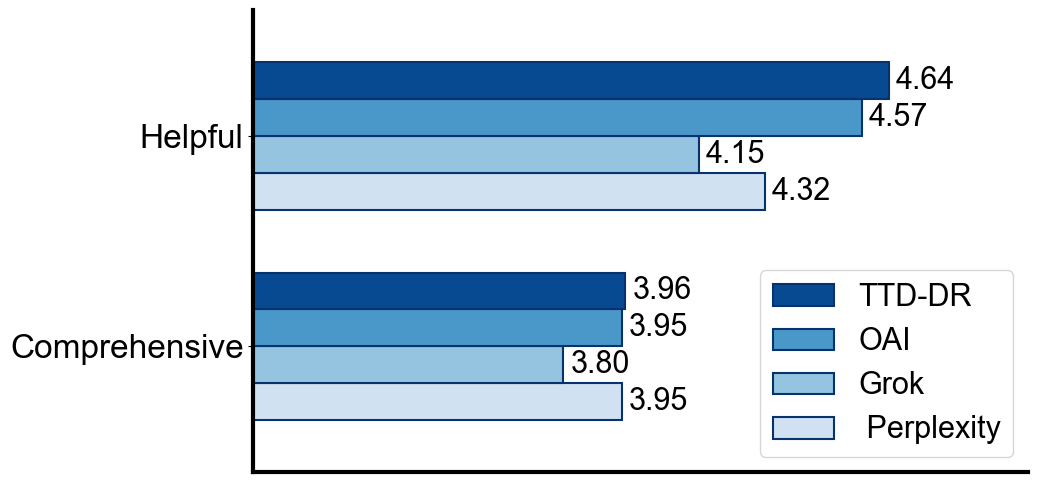}
		\caption{DeepConsult}
	\end{subfigure}
	\caption{Single-sided evaluation metrics comparisons between OpenAI Deep Research and our \textbf{TTD-DR} for \textsc{LongForm Research} (left) and \textsc{DeepConsult} (right) benchmarks. \textbf{TTD-DR}'s reports tends to be more helpful and comprehensive than other DR agents.}
	\label{fig:results-single-sided}
\end{figure}

Table~\ref{tab:results-main} presents the performance comparisons between our \textbf{TTD-DR} and other DR systems. Our \textbf{TTD-DR} consistently achieves superior results across all benchmarks. Specifically, when compared to \textsc{OpenAI Deep Research}, our method achieves 69.1\% and 74.5\% win rate in side-by-side comparisons for the two \textit{long-form} research report generation tasks. Additionally, it outperforms \textsc{OpenAI Deep Research} by 4.8\%, 7.7\% and 1.7\% on the three extensive research datasets with \textit{short-form} ground-truth answers. Figure~\ref{fig:results-single-sided} further illustrates the Helpfulness and Comprehensiveness auto-rater scores for the two \textit{long-form} research tasks, where our \textbf{TTD-DR} also surpasses \textsc{OpenAI Deep Research}, particularly for the \textsc{LongForm Research} dataset.

Table~\ref{tab:results-ablation} shows the ablation study for our DR agents. It's evident that even the most advanced LLMs with strong reasoning capabilities, such as \textsc{Gemini-2.5-flash} and \textsc{Gemini-2.5-pro}, perform poorly without any search tools. For instance, on the curated \textsc{HLE-Search} dataset, \textsc{Gemini-2.5-pro}, despite showing relatively good results on the full HLE set (20.9\%), achieves only 8.6\% accuracy. The performance of both base LLMs significantly improves when augmented with search tools, though their results remain considerably lower than \textsc{OpenAI Deep Research}.

Now, examining the three agentic DR agents, the basic DR agent shows significant improvement over LLMs with search tool but still underperforms \textsc{OpenAI Deep Research}. With the addition of the proposed \textsc{Self-evolution} algorithm, we observe that for \textsc{LongForm Research} and \textsc{DeepConsult}, our system outperforms \textsc{OpenAI Deep Research} with 60.9\% and 59.8\% win rates, respectively. The Correctness scores on the two HLE datasets also show an improvement of 1.5\% and 2.8\% against OpenAI DR, respectively, although we still underperform on GAIA by 4.4\%. Finally, incorporating \textsc{Diffusion with Retrieval} leads to substantial gains over \textsc{OpenAI Deep Research} across all benchmarks.

Furthermore, we plot the Pareto frontier of our systems to study the trade-off between latency and performances. In Figure~\ref{fig:pareto-frontier}, the x-axis represents the $log_{10}$ of seconds. The left y-axis shows our \textbf{TTD-DR}'s win rate over OpenAI DR on \textsc{LongForm Research}. The data points, from left to right, represent 
\textsc{Gemini-2.5-pro w/ search tool}, \textsc{DR-Agent-Base}, \textsc{+Self-Evolution} and \textsc{+Diffusion with Retrieval} with increasing latency. The convex shape, particularly the upward trending slope of the last two points, indicates that our two proposed algorithms provide more performance gains per unit increase in latency. This demonstrates that both denoising with retrieval and self-evolution are efficient algorithms for test-time scaling.

\begin{table*}[t]
\centering
\caption{In this Table, we show the ablation study of our DR Agent's performances across all benchmark datasets. All Win rate (\%) are computed against OpenAI Deep Research. Correctness (\%) uses LLM-as-a-judge with the standard evaluation prompt.}
\resizebox{\textwidth}{!}{
\begin{tabular}{lccccc}
\toprule[1.5pt]
& \textsc{LongForm Research} & \textsc{DeepConsult} & \textsc{HLE-Search} & \textsc{HLE-Full} & \textsc{GAIA} \\
\cmidrule{2-6}
& \textbf{Win Rate} & \textbf{Win Rate} & \textbf{Correctness} & \textbf{Correctness} & \textbf{Correctness} \\
\midrule[1pt]
\textsc{OpenAI Deep Research} & - & - & 29.1 & 26.6 & 67.4 \\
\midrule[1pt]
\multicolumn{6}{l}{\textbf{LLM w/o agentic workflow}} \\
\textsc{Gemini-2.5-flash} & 21.0 & 16.7  & \phantom{0}2.8 & 11.6 & 31.5 \\
\textsc{Gemini-2.5-flash w/ search tool} & 27.8 & 17.6 & 14.6 & 14.6 & 57.6 \\
\textsc{Gemini-2.5-pro} & 31.0 & 17.6 & \phantom{0}8.6 &  20.9 & 57.0 \\
\textsc{Gemini-2.5-pro w/ search tool} & 35.0 & 19.6 & 20.0 & 21.6 & 61.8 \\
\midrule[1pt]
\multicolumn{6}{l}{\textbf{Test-Time Diffusion Deep Researcher (ours)}} \\
\textsc{Backbone DR Agent} & 39.4 & 24.5 & 26.8 & 28.6 & 61.8 \\
\textsc{+ Self-evolution} & 60.9 & 59.8 & 30.6 & 29.4 & 63.0 \\
\textsc{+ Diffusion with Retrieval} & \textbf{69.1} & \textbf{74.5} & \textbf{33.9} & \textbf{34.3} & \textbf{69.1} \\
\bottomrule[1.5pt]
\end{tabular}
}
\label{tab:results-ablation}
\end{table*}

\subsection{Analysis}
This section provides a deeper understanding of how our two proposed methods contribute to the improvements in DR agents.

\begin{figure}[t]
\centering
	\begin{subfigure}[b]{0.42\linewidth}
	\captionsetup{justification=centering,margin=0cm}
		\centering
		\includegraphics[width=\textwidth]{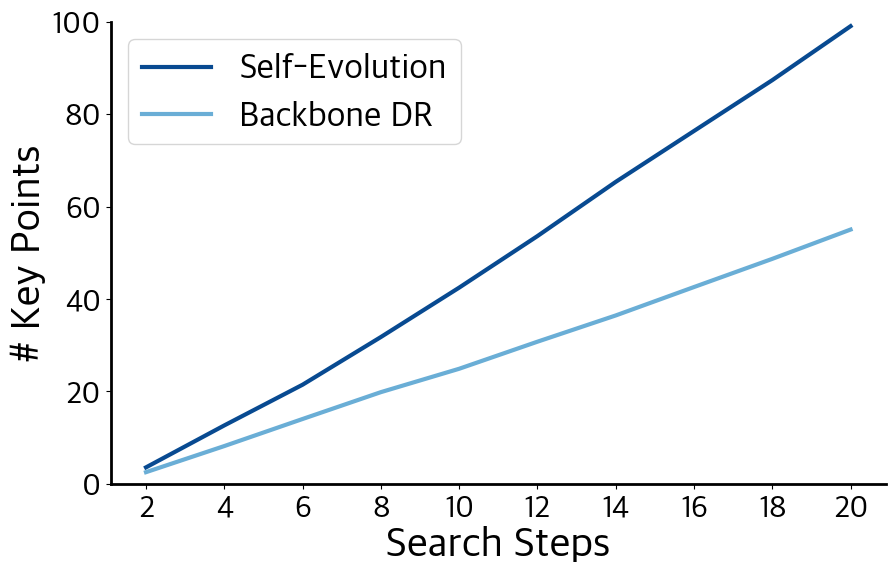}
		\caption{Query complexity comparison.}
	\end{subfigure}\qquad\qquad
	\begin{subfigure}[b]{0.42\linewidth}
	\captionsetup{justification=centering,margin=0cm}
		\centering
		\includegraphics[width=\textwidth]{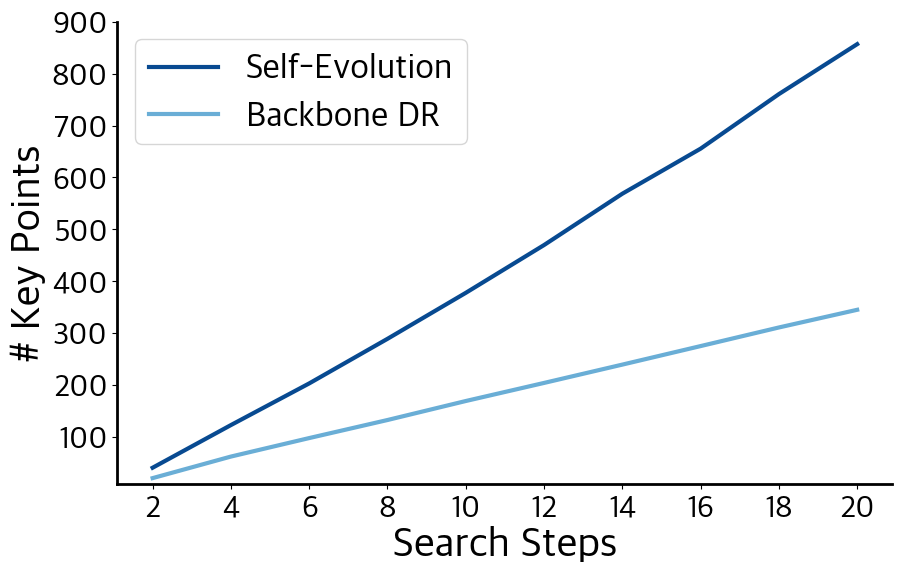}
		\caption{Answer complexity comparison.}
	\end{subfigure}
	\caption{Stage 2 generated search question (left) and answer (right) complexity by number of key points extracted by LLM using Prompt~\ref{AIbox:query-complexity} and~\ref{AIbox:answer-complexity} in the appendix. Self-evolution encourages both search question and answer diversity, which enhance the richness of information available, and thus explains the final quality improvements.}
	\label{fig:query-answer-complexity}
\end{figure}

\vspace{-3mm}
\paragraph{Improvement of self-evolution over backbone DR.} Figure~\ref{fig:query-answer-complexity} shows the cumulative complexity comparisons for search queries and answers on \textsc{DeepConsult}. Complexity is measured by key points extracted by an LLM (Gemini-2.5-pro). We observe that self-evolution significantly increases the complexity of the search process, which enriches the information gathered and, consequently, lead to better final report quality.

Our final diffusion algorithm allows for the revision and saving of intermediate reports, enabling us to assess the step-by-step report quality, as illustrated by Figure~\ref{fig:pareto-frontier-revision}. As we increase computing resources by adding more search and revision steps, we achieve increasingly significant gains against OpenAI Deep Research. Results for \textsc{HLE-Search} can be found in Appendix~\ref{sec:additional-analysis}. We next aim to understand the contributions of the denoising with retrieval algorithm to these improvements, building upon the self-evolution algorithm.

\begin{figure}[t]
\centering
	\begin{subfigure}[b]{0.3\linewidth}
	\captionsetup{justification=centering,margin=0cm}
		\centering
		\includegraphics[width=\textwidth]{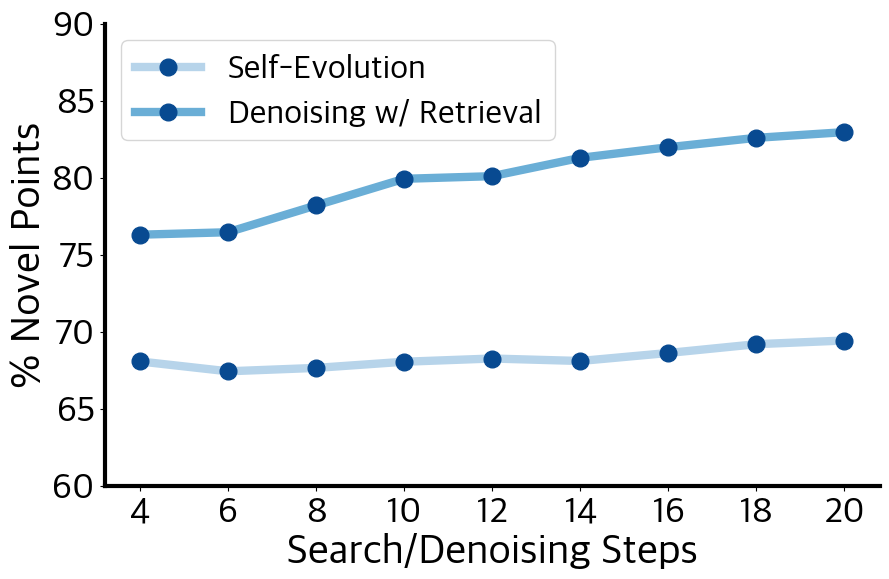}
		\caption{Cumulative search query novelty.}
		\label{fig:query-novelty}
	\end{subfigure}\qquad
	\begin{subfigure}[b]{0.3\linewidth}
	\captionsetup{justification=centering,margin=0cm}
		\centering
		\includegraphics[width=\textwidth]{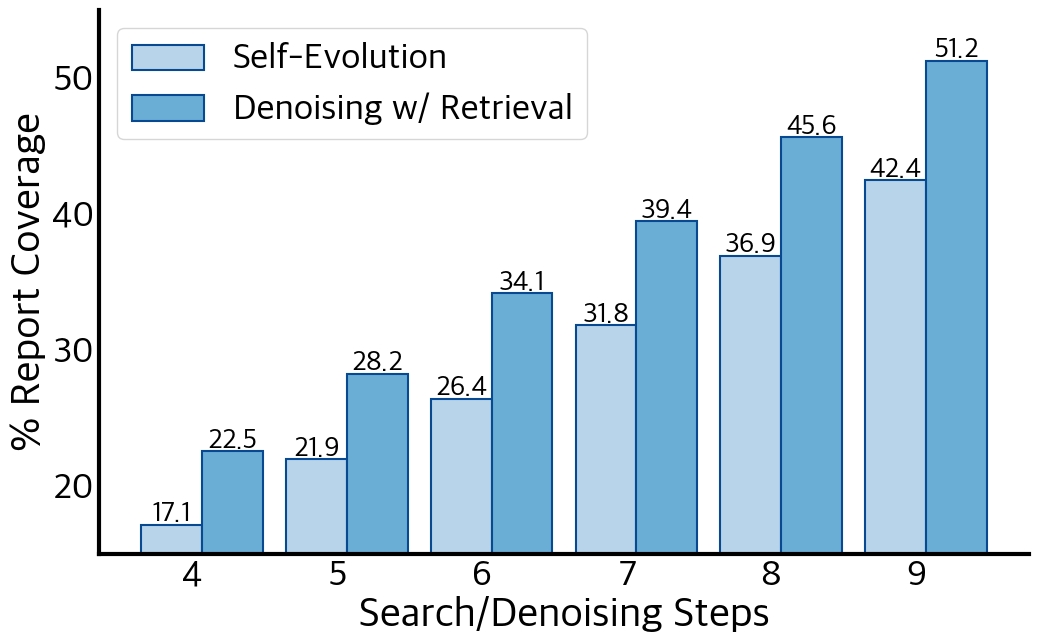}
		\caption{Report information attribution.}
		\label{fig:report-coverage}
	\end{subfigure}\qquad
	\begin{subfigure}[b]{0.3\linewidth}
	\captionsetup{justification=centering,margin=0cm}
		\centering
		\includegraphics[width=\textwidth]{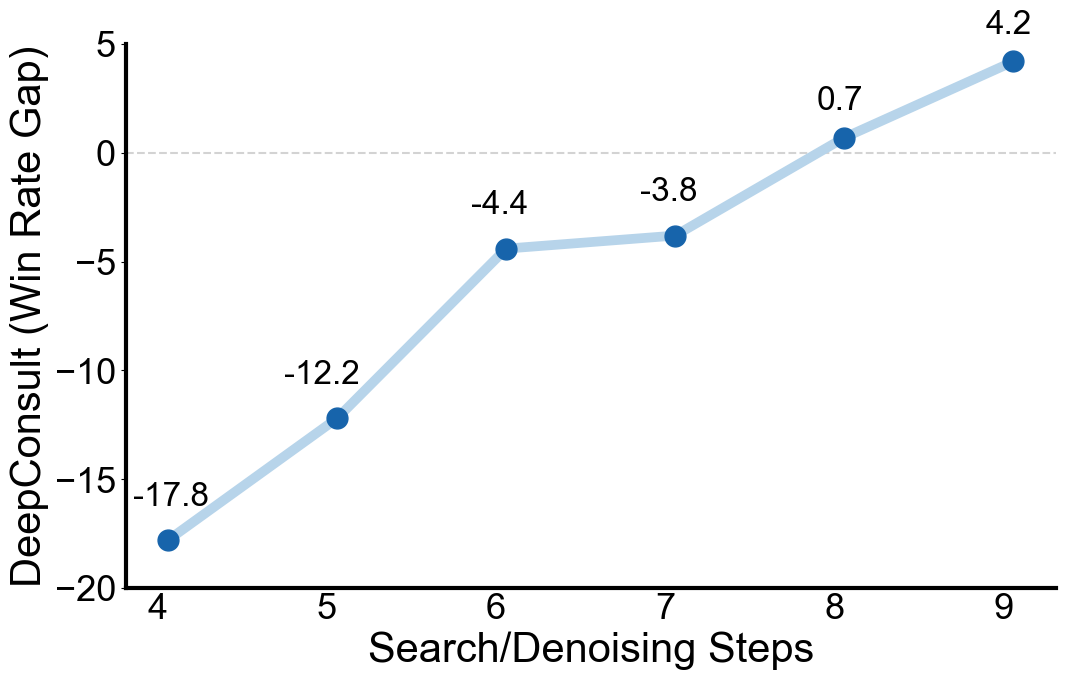}
		\caption{Performance gap: denoising v.s. self-evolution w/ 20 steps.}
		\label{fig:denoising-gap}
	\end{subfigure}
	\caption{Comparisons between denoising with retrieval and self-evolution algorithms. \textbf{(a)}: percentage of cumulative novel points (Prompt~\ref{AIbox:query-novelty}) in Stage 2 generated search queries, which shows denoising with retrieval algorithm guides the generation of more unexplored search queries. \textbf{(b)}: cumulative information attribution of the final report in Stage 2 search answers (Prompt~\ref{AIbox:report-coverage}), which demonstrates that our final method incorporates information timely in early search stages. \textbf{(c)} shows the performance gap between early steps of denoising v.s. self-evolution with full 20 search steps. With only 9 steps, denoising w/ retrieval already incorporates 51.2\% of the final report information, and outperforms self-evolution with 20 steps by 4.2\% per win ratio.}
\end{figure}

\vspace{-3mm}
\paragraph{Improvement of denoising with retrieval over self-evolution.} Figure~\ref{fig:query-novelty} displays the cumulative search query novelty comparisons on \textsc{DeepConsult}. Novelty is measured by the percentage of cumulative new points generated (extracted by Gemini-2.5-pro using Prompt~\ref{AIbox:query-novelty}). We can observe that denoising with retrieval increases query novelty by more than \textbf{12 percentage points} throughout the search and revision process by feeding the revised report to guide the exploration of new queries. In Figure~\ref{fig:report-coverage}, we present the report attribution in answers (computed using Gemini-2.5-pro with Prompt~\ref{AIbox:report-coverage}) during early search and revision steps. Notably, at Step 9, denoising w/ retrieval already incorporates \textbf{51.2\% of the final report information}, and outperforms self-evolution (with 20 search steps) by \textbf{4.2\% in win ratio} (last point in Figure~\ref{fig:denoising-gap}). These results indicate that denoising with retrieval effectively leverage information in early stages, leading to timely preservation of knowledge when agents are learning most efficiently, as shown in Figure~\ref{fig:pareto-frontier-revision}.

\section{Related Work}
We review related work that motivates our deep research agents.

\noindent
\textbf{Test-time compute scaling.} \citet{baek2024researchagent, lu2024aiscientist, zheng-etal-2024-openresearcher} are earlier efforts to build research assistant/scientist agents with search tools and iterative refinement algorithms during test time. More recently, \citet{ai-coscientist} proposes an AI Co-scientist agent for biomedical research integrating test-time algorithms such as debates mechanism to generate novel ideas, tournaments to compare and rank research hypothesis and self-critique to refine research proposals. \citet{schmidgall2025agentlaboratoryusingllm} builds an end-to-end scientific paper writing agent with self-reflection at each stage of their agent workflow. Notably, they enable a co-pilot mode where a human can step in and provide feedback, which is shown to improve overall paper quality. \cite{ai-scientist-v2} designs a machine learning research agent by incorporating a tree-search inference algorithm that is able to write a full research paper accepted by ICLR workshop. \cite{airesearcher} proposes a multi-agent system that is able to review literature, generate new ideas, invent new algorithms, conduct experiments and draft a publication-ready paper. Similarly, \cite{deerflow} leverages a multi-agent system with planner, coordinator, researcher and reporter to produce comprehensive responses to general user queries.

Amongst test-time algorithms, self-evolving \citep{lee-evolving, alpha-evolve, alita} emerges recently as a popular framework to design various agentic systems including DR. Our \textbf{self-evolution} algorithm shares common spirit with this method, particularly in its capability to conduct multiple self-critique and self-refinements. However, \textbf{TTD-DR} differs from self-evolving in that 1) our framework is fundamentally driven by human cognitive behavior, and we draw the commonality between retrieval augmented diffusion process and human writing process to develop our test-time diffusion DR; 2) Self-evolution improves individual agents to provide high quality contextual information to assist the main denoising algorithm. Both human cognitive behavior and the interplay of self-evolution and denoising with retrieval are not explicitly modeled in prior work.

\noindent
\textbf{Agent Tuning.} A few recent works explored improving deep research agent via training. Earlier work focuses on building an agentic RAG system that is able to conduct deep search and reasoning. \cite{guan2024amorrecipebuildingadaptable} proposes a multitask learning objective with both component-wise SFT data and model feedback to jointly train each module in its agentic RAG system. \citet{jin2025search} converts search actions and LLM final responses into a single sequence input, and train the RAG system end-to-end with final response reward. More recently, \citet{webthinker}, \citet{zhengdeep}, \citet{pangu}, and \citet{kimi-researcher} leverage reinforcement learning to training a research assistant agent that is able to leverage search and browsing tools to collect information and write reports. In our work, we focus on test-time compute, and leave agent tuning for future work.

\noindent
\textbf{LLM diffusion models.} Traditional LLM training paradigm leverages autoregressive objective to train models and sample outputs. LLM Diffusion models attempt to improve the scalability of state-of-the-art LLMs by breaking the assumption of sampling from first to the last tokens. LLM diffusion models are trained to first generate a complete "noisy" draft, and they iteratively denoise multiple tokens into a full high quality draft \citep{diffusionsurvey, llm-diffusion, gemini-diffusion}. Due to highly parallelizable generation processing, this line of work has the potential to achieve higher efficiency while preserving quality. Our work is inspired by LLM Diffusion models by introducing the denoising mechanism during test-time report writing, but differ from them in that we do not train our agents; instead we assume LLM agents are well crafted to perform denoising tasks.
\section{Conclusions}
The Deep Researcher with Test-Time Diffusion (TTD-DR) agent is a novel framework for generating research reports, inspired by the iterative nature of human research. This agent addresses the limitations of existing DR agents by conceptualizing report generation as a \textit{diffusion process}. TTD-DR initiates with a preliminary draft, an updatable skeleton that guides the research direction. This draft is then refined iteratively through a ``denoising'', dynamically informed by a retrieval mechanism that incorporates external information at each step. The core process is further enhanced by a self-evolutionary algorithm applied to each component of the agentic workflow, ensuring the generation of high-quality context for the diffusion process.

The TTD-DR framework achieves state-of-the-art results across various benchmarks requiring intensive search and multi-hop reasoning, significantly outperforming existing DR agents. It demonstrates superior performance in generating comprehensive long-form research reports and identifying concise answers for multi-hop search and reasoning tasks. The framework's draft-centric design guides the report writing process to be more timely and coherent while reducing information loss during the iterative search process.

\section*{Limitations}
While TTD-DR shows significant advancements, the current work primarily focuses on search tool usage and does not incorporate other tools such as browse and coding. Future work could explore integrating these additional tools to further enhance the DR agents' performance and broaden their application. Additionally, agent tuning for improving deep research agents is left for future work, as the current focus is on test-time scaling.

\bibliographystyle{abbrvnat}
\nobibliography*
\bibliography{custom}

\clearpage
\appendix
\section{Appendix}
\label{sec:appendix}

\subsection{Evaluation Guidelines}
\label{sec:eval-guideline}
\paragraph{Helpfulness} categories can be found below.
\begin{itemize}
    \item Very Helpful: all statements are helpful based on the guideline above.
    \item Helpful: Most statements are helpful except for 1-2 statements with minor issues according to the guideline above.
    \item Mostly Helpful: 1-2 statements seriously fail the guideline above, or 3-5 statements have minor issues.
    \item Somewhat Helpful: > 2 statements with serious issues, or > 5 statements with minor issues.
    \item Not at all Helpful: None statements are helpful.
\end{itemize}

\paragraph{Comprehensiveness} categories can be found below.
\begin{itemize}
    \item Very Comprehensive: it is hard to identify any points that could be added to the report to make it more comprehensive.
    \item Comprehensive: it is hard to identify any major points that could be added to the report to make it more comprehensive. It would be nicer to add some minor points, but not necessary. 
    \item Mostly Comprehensive: There 1-2 major points that should be added to the report.
    \item Somewhat Comprehensive: There are more than 2 major points that should be added.
    \item Not at all Comprehensive: There are more than 5 major points that should be added.
\end{itemize}

\subsection{Human Annotation Interface}
\begin{figure*}[t]
    \centering
\includegraphics[trim={0cm 0cm 0cm 0cm}, clip, width=\columnwidth]{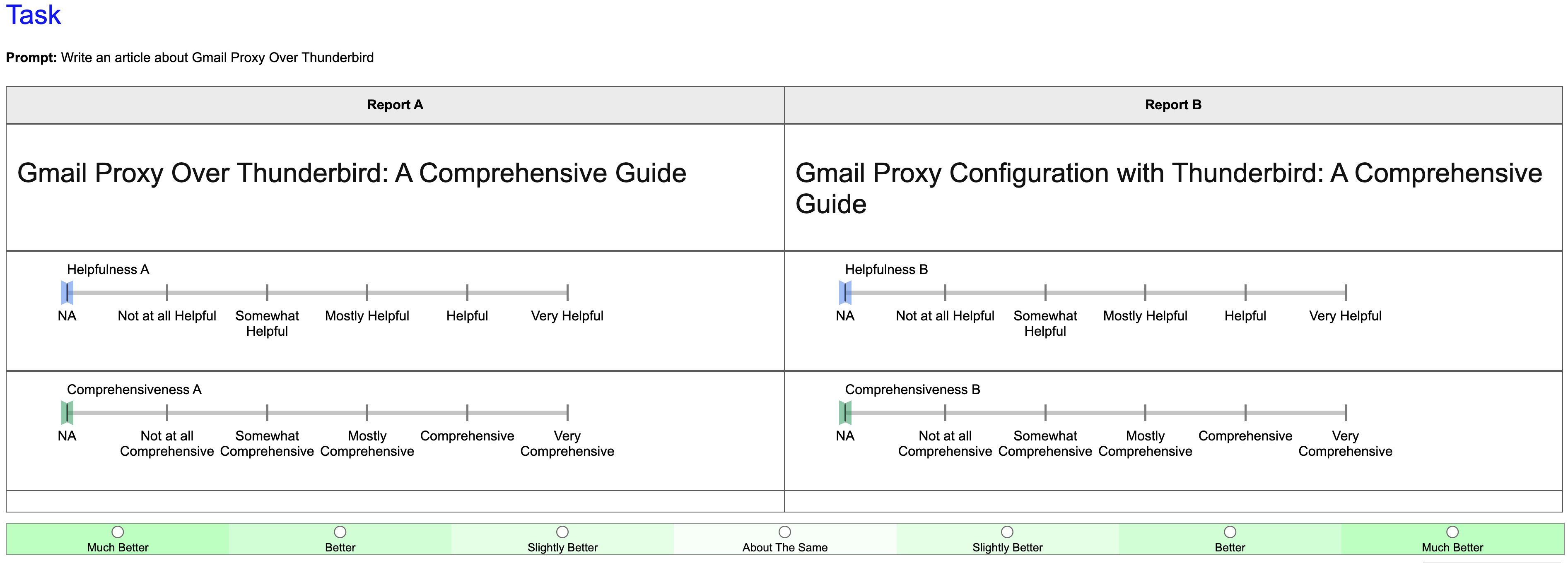}
\caption{Helpfulness, Comprehensiveness, and side-by-side rating between Report A and B. Report are simplified for clarify purpose.}
\label{fig:interface}
\end{figure*}

Figure~\ref{fig:interface} shows our human annotation interface.
\label{sec:eval-interface}

\subsection{Human and LLM-as-a-judge Alignment}
\label{sec:human-llm-align}
\begin{table}
\centering
\caption{In this Table, we show the alignments between our auto-rater and human raters. Human accuracy is computed comparing two raters' scores by treating one as ground-truth and taking average.}
\resizebox{0.55\textwidth}{!}{
\begin{tabular}{lcc}\toprule[1.5pt]
Evaluator Models & Correlation & 
Accuracy \\
\midrule
\textsc{Gemini-1.5-pro-002} & 0.22 & 60.8 \\
\textsc{Gemini-2.0-flash-001} & 0.07 & 51.1 \\
\textsc{Gemini-2.5-pro-preview-03-25} & 0.12 & 47.8 \\
\midrule
\textsc{Human} & - & 69.0 \\
\bottomrule
\end{tabular}
}
\label{tab:results-align}
\end{table}

\subsection{HLE Query Categorization}
\label{sec:query-cat}
We use the following prompt to categorize HLE queries into 1) reasoning only and 2) reasoning+search.

\begin{AIbox}{HLE Query Categorization Prompt}
You are an expert categorizing a query from a user. Your task is to assign the query to one of the following 2 categories:\\
* "Reasoning": The query can be answered with pure logical reasoning without any external world knowledge.\\
* "Search": The query can NOT be answered with pure logical reasoning, but requires additional information that can be obtained through searching the web. \\

The query is in the <query></query> tags and the answer to the query is in <reference></reference>. \\
We also provide rational in <rational></rational> that explains the answer. \\

First, follow the instructions in the <instructions></instructions> tags below to assess the Correctness of the answer. \\

<rubrics>\\
Please output using the scale below: \\
* 1: Reasoning: The query can be answered with pure logical reasoning without any external world knowledge.\\
* 2: Search: The query can NOT be answered with pure logical reasoning, but requires additional information that can be obtained through searching the web.\\
</rubrics>\\

Here is the query:\\
<query>\\
\{query\}\\
</query>\\

Here is the answer:\\
<reference>\\
\{answer\}\\
</reference>\\

Here is the rational that leads to the reference answer:\\
<rational>\\
\{rational\}\\
</rational>\\

Review the rubrics in the <rubrics></rubrics> tags above to rate the answer.\\
First, think step by step, put your thinking in <thinking></thinking> tags. Your thinking must be shorter than 200 words. Then, provide your category inside <rating></rating> tags. Remember your output must be either 1 or 2 in <rating></rating> tags.
\label{AIbox:hle-query-cat}
\end{AIbox}

\subsection{Answer Merging.}
\label{sec:report-merge}
We use the following prompt to merge multiple answer into one for the parallel denoising algorithm.

\begin{AIbox}{Answer Merging Prompt}
Your task is to research a topic and try to fulfill the user query in the <user> tags. \\

<instructions> \\
You are given a list of candidate answers in <answer\_list> tags below. Combine them into a single answer so that,\\
+ it best fulfills the initial user query in the <user> tags. \\
+ If there are conflicting information, try to reconcile them in a logically sound way. \\
</instructions> \\

Here is the user query. \\
<user> \\
\{query\} \\
</user> \\

Here is the list of candidate answers you need to merge. \\
<answer\_list> \\
\{answer\_list\} \\
</answer\_list> \\

Only output a combined answer from the answers in <answer\_list>. Do NOT use other information. \\
\label{AIbox:report-merge}
\end{AIbox}

\subsection{Hyper-parameters}
\label{sec:hyper}
We list a few key hyper-parameters for our self-evolution algorithm shown in Fig.~\ref{fig:parallel-diffuse}. To recap, this algorithm generates multiple initial states, each undergoes self-evolving steps before being merged into a final one. So it introduces two sets of hyper-parameters: $n$ number of initial states and $s$ number of evolving steps.
\begin{table}[h]
\centering
\resizebox{0.98\textwidth}{!}{
\begin{tabular}{llcccc}\toprule[1.5pt]
Hyper-parameters & Description & \textsc{LongForm Research} & \textsc{DeepConsult} & \textsc{HLE} & \textsc{GAIA} \\
\midrule
$n_p$ & No. of initial plan states & 1 & 1 & 1 & 1 \\
$n_{q}$ & No. of initial search query states & 5 & 5 & 5 & 5 \\
$n_{a}$ & No. of initial answer states & 3 & 3 & 3 & 3 \\
$n_{r}$ & No. of initial report states & 1 & 1 & 5 & 5 \\
$s_{p}$ & No. of plan self-evolving steps & 1 & 1 & 1 & 1 \\
$s_{q}$ & No. of search query self-evolving steps & 0 & 0 & 0 & 0 \\
$s_{a}$ & No. of answer self-evolving steps & 0 & 0 & 0 & 0\\
$s_{r}$ & No. of report self-evolving steps & 1 & 1 & 0 & 0 \\
\bottomrule
\end{tabular}
}
\caption{We show hyper-parameter description and best settings in this table.}
\label{tab:hyper}
\end{table}

\subsection{Question Complexity}
\begin{AIbox}{Unique Question Key Points Extraction}
You are provided with a question in <question> tag. Analyze the complexity of the question. \\

<question> \\
\{question\} \\
</question> \\
\\
Breakdown the question into unique key points, and then calculate the number of key points in the question. \\

First, put your thinking in <thinking></thinking> tags, and then put the number in <number></number> tags. \\
Return an integer. \\
\label{AIbox:query-complexity}
\end{AIbox}

\subsection{Answer Complexity}
\begin{AIbox}{Unique Answer Key Points Extraction}
You are provided with an answer in <answer> tag. Analyze the complexity of the answer. \\

<answer> \\
\{answer\} \\
</answer> \\

Breakdown the answer into unique key points, and then calculate the number of key points in the answer. \\

First, put your thinking in <thinking></thinking> tags, and then put the number in <number></number> tags. \\
Return an integer. \\
\label{AIbox:answer-complexity}
\end{AIbox}

\subsection{Query Novelty}
\begin{AIbox}{Search Question Novelty}
You are provided with a list of used questions in <question\_list> tags and a new question in <new\_question> tags. \\
You need to judge how novel the new question is given the used questions. \\
\\
<question\_list> \\
\{question\_list\} \\
</question\_list>\\
\\
<new\_question>\\
\{new\_question\} \\
</new\_question> \\
\\
Breakdown the new question into unique key points, and then calculate the number of key points that are NOT semantically covered in any of the used questions. \\

First, put your thinking in <thinking></thinking> tags, and then put the number in <number></number> tags. \\
Return an integer. \\
\label{AIbox:query-novelty}
\end{AIbox}

\subsection{Report Coverage}
\begin{AIbox}{Report Coverage}
Given a context in <context> tags, you need to judge how much content in this context is included in the response in <response> tags. \\

<context> \\
\{context\} \\
</context> \\

<response> \\
\{response\} \\
</response> \\

Breakdown the context into sentences, and then calculate the ratio of sentences that are semantically covered in response. \\

First, put your thinking in <thinking></thinking> tags, and then put the ratio in <ratio></ratio> tags.
Round the ratio to 2 decimal places. \\
\label{AIbox:report-coverage}
\end{AIbox}

\subsection{Additional Analysis Results}
\label{sec:additional-analysis}
\begin{figure}
\centering
	\begin{subfigure}[a]{0.5\linewidth}
	    \captionsetup{justification=centering,margin=0cm}
		\centering
		\includegraphics[width=0.97\textwidth]{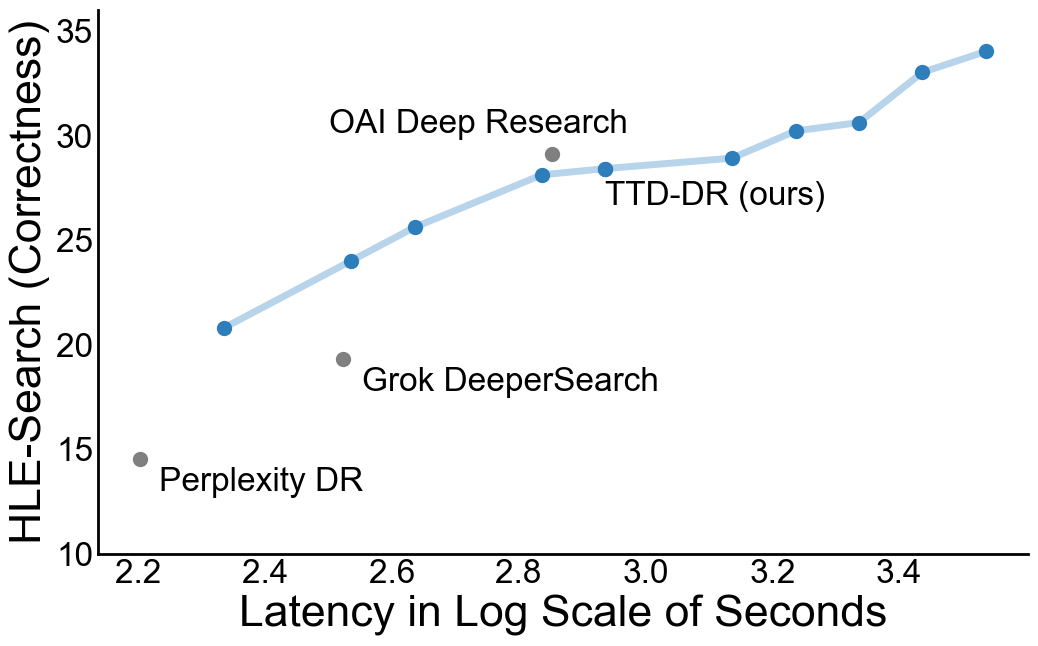}
	   \caption{Pareto frontier for different DR designs.}
	    \label{fig:pareto-frontier-revision-hle}
	\end{subfigure}
	\caption{Pareto frontier between DR agent performances and latency for \textsc{HLE-Search}. The dots from left to right represent adding more search/revision steps up to 20, which shows with similar latency, we achieve on-par or better results compared with competing DR agents. Note that HLE dataset only requires identify short-form answer, which does not align perfectly well with our primary tasks of writing real-world long-form reports.}
\end{figure}

\begin{figure}
\centering
	\begin{subfigure}[a]{0.47\linewidth}
	    \captionsetup{justification=centering,margin=0cm}
		\centering
		\includegraphics[width=0.97\textwidth]{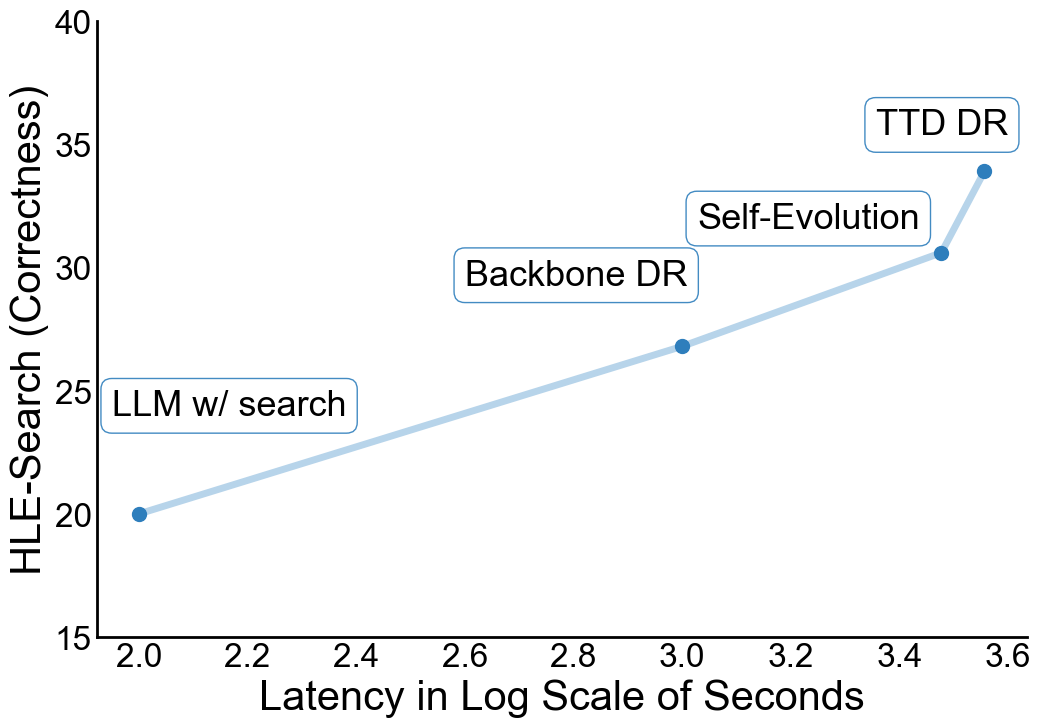}
	    \label{fig:pareto-frontier-hle}
	\end{subfigure}
	\caption{Pareto frontier between DR agent performances and latency for HLE-search. The dots from left to right represent 1) \textsc{Gemini-2.5-pro w/ search tool}, 2) \textsc{Backbone DR Agent}, 3) \textsc{+Self-evolution} and 4) \textsc{+Diffusion with Retrieval}, which shows our final algorithm is most efficient in terms of test-time scaling (steepest slope).}
\end{figure}

\end{document}